\documentclass[journal]{IEEEtran}
\IEEEoverridecommandlockouts
\usepackage{cite}
\usepackage{amsmath,amssymb,amsfonts}
\usepackage{algorithmic}
\usepackage{graphicx}
\usepackage{textcomp}
\usepackage{subcaption}
\usepackage{xcolor}
\usepackage{booktabs,cite}
\newcommand\mydots{\hbox to 1em{.\hss.\hss.}}
\usepackage{soul}
\usepackage{bbold}
\usepackage{amsthm}
\newtheorem*{theorem*}{Theorem}
\newtheorem*{lemma*}{Lemma}
 
\newtheorem{assumption}{Assumption}  
\newtheorem{definition}{Definition}  

\newcommand{\blue}[1]{\textcolor{black}{#1}}

\DeclareMathOperator*{\argmax}{arg\,max}
\DeclareMathOperator*{\argmin}{arg\,min}
\DeclareMathOperator*{\minimize}{minimize}

\newcommand{\norm}[1]{\left\lVert#1\right\rVert}
\def\BibTeX{{\rm B\kern-.05em{\sc i\kern-.025em b}\kern-.08em
    T\kern-.1667em\lower.7ex\hbox{E}\kern-.125emX}}
\begin{document}

\title{Robust Bayesian Learning for Reliable Wireless AI: Framework and Applications
\thanks{The work of M. Zecchin and D.Gesbert is funded by the Marie Curie action WINDMILL (grant No. 813999), while O. Simeone and S. Park have received funding from the European Research Council (ERC) under the European Union’s Horizon 2020 Research and Innovation Programme (Grant Agreement No. 725731). M. Kountouris has received funding from the European Research Council (ERC) under the European Union’s Horizon 2020 research and innovation programme (Grant agreement No. 101003431).

Matteo Zecchin, Marios Kountouris and David Gesbert are with the Communication Systems Department, EURECOM, Sophia-Antipolis, France (e-mail: zecchin@eurecom.fr; kountour@eurecom.fr; gesbert@eurecom.fr).

    Sangwoo Park and Osvaldo Simeone are with the King's Communications, Learning \& Information Processing (KCLIP) lab, Department of Engineering, King’s College London, London WC2R 2LS, U.K. (e-mail: sangwoo.park@kcl.ac.uk; osvaldo.simeone@kcl.ac.uk).

The second author has contributed to the problem definitions and to the experiments. The third author has had an active role in defining the problems, as well as in writing the text, while the last two authors have had a supervisory role.
}
}

\author{Matteo Zecchin,~\IEEEmembership{Student Member,~IEEE,} Sangwoo Park,~\IEEEmembership{Member,~IEEE,} Osvaldo Simeone,~\IEEEmembership{Fellow,~IEEE,} Marios Kountouris,~\IEEEmembership{Senior Member,~IEEE}, David Gesbert,~\IEEEmembership{Fellow,~IEEE} 
}
\maketitle

\begin{abstract}
This work takes a critical look at the application of conventional machine learning methods to wireless communication problems through the lens of reliability and robustness. Deep learning techniques adopt a frequentist framework, and are known to provide poorly calibrated decisions that do not reproduce the true  uncertainty caused by limitations in the size of the training data. Bayesian learning, while in principle capable of addressing this shortcoming, is in practice impaired by model misspecification and by the presence of outliers. Both problems are pervasive in wireless communication settings, in which the capacity of machine learning models is subject to resource constraints and training data is affected by noise and interference.  In this context, we explore the application of the framework of \emph{robust} Bayesian learning. After a tutorial-style introduction to robust Bayesian learning, we showcase the merits of robust Bayesian learning on several important wireless communication problems in terms of accuracy, calibration, and robustness to outliers and misspecification.
\end{abstract}


\section{Introduction}


Artificial intelligence (AI) is widely viewed as a key enabler of 6G wireless systems. 
Research on this topic has mostly focused on identifying use cases and on mapping techniques from the vast literature on machine learning to given problems \cite{simeone2018very,sun2019application,gunduz2019machine}. At a more fundamental level, there have been efforts to integrate well-established communication modules, e.g., for channel encoding and decoding, with data-driven designs, notably via tools such as model unrolling \cite{jiang2020learn,kim2018deepcode}. All these efforts have largely relied on \emph{deep learning} libraries and tools. The present paper takes a critical look at the use of this conventional methodology through the lens of \emph{reliability} and \emph{robustness} . To this end, we explore the potential benefits of the alternative design framework of \emph{robust Bayesian learning} by focusing on several key wireless communication applications, namely modulation classification, indoor and outdoor localization, and channel modeling and simulation. 

\subsection{Frequentist vs. Bayesian Learning}

In \emph{frequentist} learning, the output of the training process is a single model -- typically, a single vector of weights for a neural network -- obtained by minimizing the training loss. This approach is justified by the use of the training loss as an estimate of the population loss, whose computation would require averaging over the true, unknown distribution of the data. This estimate is only accurate in the presence of sufficiently large data sets. While abundant data is common in the benchmark tasks studied in the computers science literature, the reality of many engineering applications is that data are often scarce. In wireless communications, the problem is particularly pronounced at the physical layer, in which fading dynamics imply short stationary intervals for data collection and training \cite{park2020meta,park2020learning,simeone2020learning,yuan2020transfer}.

The practical upshot of the reliance on frequentist learning is that, in the presence of limited data, decisions made by AI models tend to be \emph{poorly calibrated}, providing confidence levels that do not match their true accuracy  \cite{guo2017calibration,cohen2021learning}. As a result, an AI model may output a decision with some level of confidence, say $95\%$, while the accuracy of the decision is significantly lower. This is an issue problem in many engineering applications, including emerging communication networks (e.g., 5G and beyond), in which a more or less confident decision should be treated differently by the end user \cite{masur2021artificial}. 

The framework of \emph{Bayesian learning} addresses the outlined shortcomings of frequentist learning \cite{mackay2003information,osawa2019practical}. At its core, Bayesian learning optimizes over a \emph{distribution} over the model parameter space, which enables it to quantify uncertainty arising from limited data. In fact, if several models fit the data almost equally well, Bayesian learning does not merely select one of the models, disregarding uncertainty; rather it assigns similar distribution values to all such models \cite{nikoloska2021bamld}. This way, decisions produced by AI modules trained via Bayesian learning can account for the ``opinions'' of multiple models by averaging their outputs using the optimized distribution  \cite{madigan1996bayesian,osvaldo2022ML4ENG}. Bayesian learning has recently been applied in \cite{cohen2021learning} by focusing on the problem of demodulation over fading channels; as well as in \cite{zilberstein2022annealed} for detection over multiple-antenna channels.

\subsection{Robust Bayesian Learning}

Like frequentist learning, Bayesian learning assumes that the distribution underlying training data generation is the same as that producing test data. Furthermore, Bayesian learning implicitly assumes that the posited model -- namely likelihood and prior distribution -- is sufficiently close to the true, unknown data-generating distribution to justify the use of the posterior distribution as the optimized distribution in the model parameter space. As a result, the benefit of Bayesian learning is degraded when data is affected by outliers and/or when the model is misspecified.

Recent work has addressed both of these limitations, introducing a generalized framework that we will refer to  as \emph{robust Bayesian learning}. Robust Bayesian learning aims at providing well-calibrated, and hence reliable, decisions even in the presence of model misspecification and of discrepancies between training and testing conditions. 

Model misspecification has been addressed in \cite{masegosa2020learning,morningstar2020pac}. These papers start from two observations. The first is that Bayesian learning can be formulated as the minimization of a \emph{free energy} metric, which involves the average of the training loss, as well as an information-theoretic regularizing term dependent on a prior distribution. The conventional free energy metric can be formally derived as an upper bound on the population loss within the theoretical framework of \emph{PAC Bayes theory} \cite{jose2021free,catoni2003pac,alquier2021user}. The second observation is that, in the presence of model misspecification, \emph{model ensembling} can be useful in combining the decisions of different models that may be specialized to distinct parts of the problem space. Using these two observations, references \cite{masegosa2020learning,morningstar2020pac} introduced alternative free energy criteria that are based on a tighter bound of the population loss for ensemble predictors.

To address the problem of outliers (see e.g. \cite{knoblauch2019generalized}), different free energy criteria have been introduced, which are less sensitive to the presence of outliers. These metrics are based on divergences, such as $\beta$-divergences \cite{basu1998robust,ghosh2016robust} and $\gamma$-divergence \cite{fujisawa2008robust,nakagawa2020robust}, which generalize the Kullback-Liebler divergence underlying the standard free energy metric.  Finally, a unified framework has been introduced in \cite{zecchin2022robust} that generalizes the free energy metrics introduced in \cite{masegosa2020learning,morningstar2020pac}. The approach is robust to misspecification, while also addressing the presence of outliers.

\subsection{Main Contributions}
In this paper, we explore the application of robust Bayesian learning to wireless communication systems. Our main purpose is twofold. On the one hand, we present a tutorial-style review of robust Bayesian learning in order to introduce the framework for an audience of communication engineers. On the other hand, we detail applications of robust Bayesian learning to communication systems, focusing on automated modulation classification (AMC), received signal strength indicator (RSSI)-based localization, as well as channel modeling and simulation.  These applications have been selected in order to highlight the importance of considering uncertainty quantification, in addition to accuracy, while also emphasizing the problems of model misspecification and outliers in wireless communications \cite{fawzy2013outliers,jin2015rssi,kalyani2007ofdm}. 

Our specific contributions are as follows. \begin{itemize}
\item We give a self-contained introduction to Bayesian and robust Bayesian learning by describing conceptual underpinnings and practical implications. 
\item We detail a series of applications of robust Bayesian learning to popular wireless communication problems, which are characterized by model misspecification and for which training must contend with data sets corrupted by outliers. 
\item As a first application, we focus on the AMC problem for intelligent spectrum sensing \cite{liang2011cognitive}. In this setting, the necessity of deploying lightweight models that satisfy the strict computational requirements of network edge devices can give rise to model misspecification. At the same time, the training data sets often contain non-informative outliers due to interfering transmissions from other devices. We demonstrate that robust Bayesian learning yields classifiers with good calibration performance despite model misspecification and the presence of outliers.
\item As a second application, we study node localization based on crowdsourced RSSI data sets \cite{lohan2017wi}. Such data sets typically contain inaccurately reported location measurements due to imprecise or malicious devices. Furthermore, owing to the complex relation between RSSI measurements and device locations, learning often happens using misspecified model classes. In this context, we demonstrate that robust Bayesian is able to properly estimate residual uncertainty about the transmitters' locations in spite of the presence of outliers and misspecified model classes. 
\item Finally, we apply robust Bayesian learning to the problem of channel modeling and simulation. We show via experiments that robust Bayesian learning produces accurate and well-calibrated generative models even in the presence of outlying data points.
\end{itemize}

\subsection{Organization}
This paper contains two main parts. In the first part, consisting of Sections \ref{sec:FvB} and \ref{sec:RobBayes}, we provide a tutorial-style review of robust Bayesian learning, along with the necessary background. The second part of the paper elaborates on the application discussed in the previous subsection.

We start the first part in Section \ref{sec:FvB}, where we define the learning setup and we provide a tutorial-style comparison between frequentist and Bayesian learning frameworks. In Section \ref{sec:mis}, we introduce the concept of model misspecification and we review the $m$-free energy criterion \cite{morningstar2020pac} as a tool to mitigate the effect of  misspecified model classes. In Section \ref{sec:rob_div}, we define outliers and illustrate the role of robust losses to reduce the influence of outlying data samples. Finally, in Section \ref{sec:robpacm}, we describe the robust Bayesian framework and review the robust $m$-free energy learning objective \cite{zecchin2022robust}. This approach simultaneously addresses model misspecification and outliers.

In the second part of this paper, we turn to a series of applications of robust Bayesian learning to important wireless communication problems. In Section \ref{exp:amc}, we consider the AMC task; Section \ref{exp:RSSI} studies the problem of robust RSSI-based localization; and Section \ref{exp:ch_sim} focuses on channel modeling and simulation. Finally, Section \ref{sec:concs} concludes the paper.
\section{Frequentist vs. Bayesian Learning}
\label{sec:FvB}

\begin{figure}
   \includegraphics[width=1\linewidth]{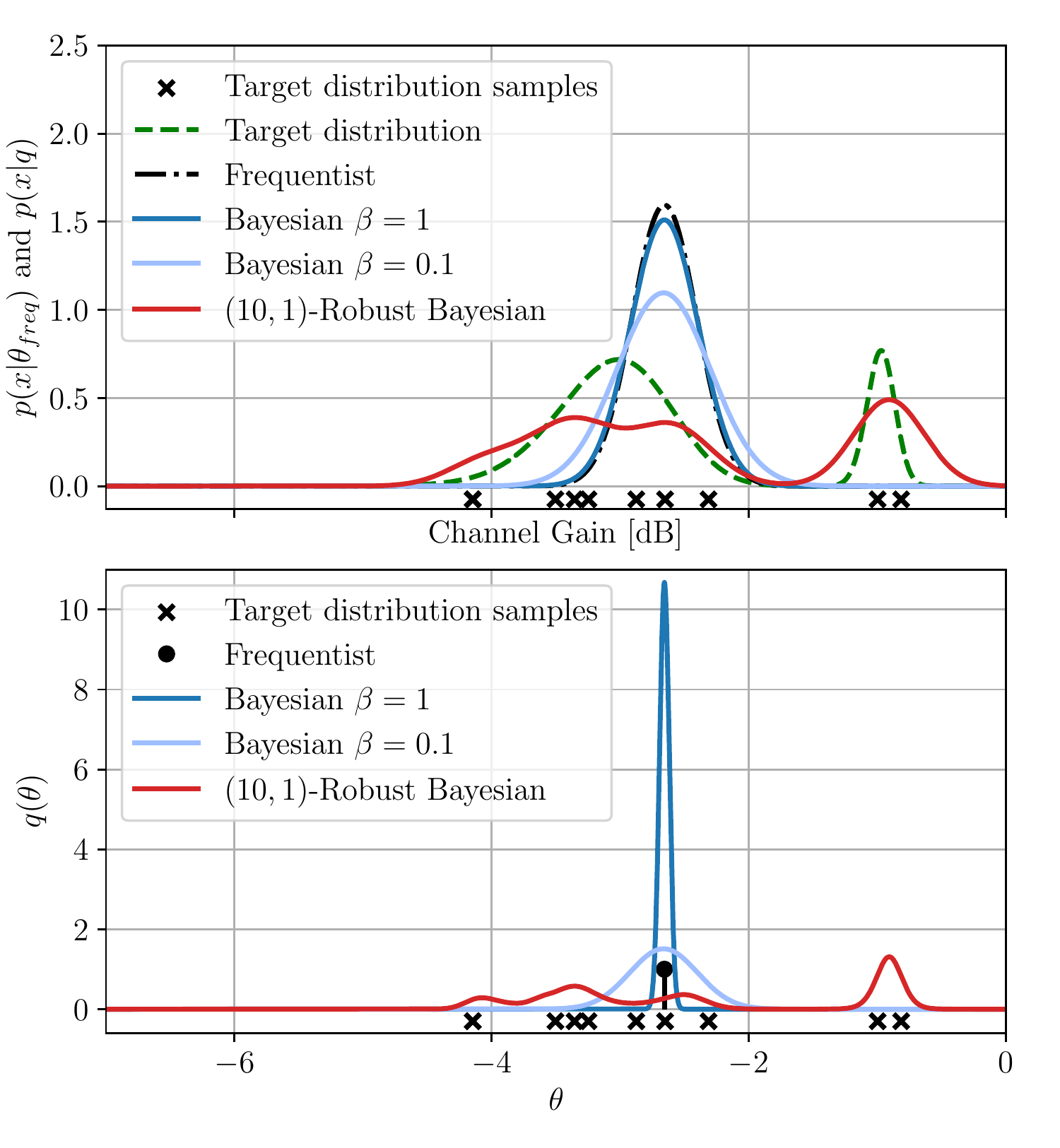}
    \caption{Estimated distribution over a scalar channel gain (top panel) and corresponding posterior distribution $q(\theta)$ over the model parameter $\theta$ (bottom panel) for frequentist learning, Bayesian learning with $\beta\in\{1,0.1\}$ and $(m,1)$-robust Bayesian learning with $m=10$. The training data set, represented  as crosses, is sampled from the target distribution $\nu(x)$.} 
    \label{fig:toy_example}
\end{figure}
\label{Sec:learning_set_up}
Throughout this paper we consider a standard learning set-up in which the learner has access to a data set $\mathcal{D}$ of $n$ data points $\{z_i\}^n_{i=1}$ sampled in an independent and identically distributed (i.i.d.) fashion from a \emph{sampling distribution} $\nu_s(z)$. As we will see, owing to the presence of outliers, the sampling distribution may differ from the \emph{target distribution} $\nu(z)$. The general goal of learning is that of optimizing models that perform well on average with respect to the target distribution $\nu(z)$. In this section, we assume that the sampling distribution $\nu_s(z)$ equals the target distribution $\nu(z)$, and we will address the problem of outliers -- which arises when $\nu_s(z) \neq \nu(z)$ -- in the next section.

We will consider both \emph{supervised learning} problems and the \emph{unsupervised learning} problem of density estimation with applications to wireless communications. In supervised learning, a data sample $z\in\mathcal{Z}$ corresponds to a pair $z=(x,y)$ that  comprises a feature vector $x\in\mathcal{X}$ and a label $y\in\mathcal{Y}$. In contrast, for density estimation,  each data point $z\in\mathcal{Z}$ corresponds to a feature vector $z=x\in\mathcal{X}$.

Supervised learning is formulated as an optimization  over a family of discriminative models defined by a parameterized conditional distribution $p(y|x,\theta)$ of target $y$ given input $x$. The conditional distribution, or model, $p(y|x,\theta)$ is  parameterized by vector $\theta\in\Theta$ in some domain $\Theta$. In contrast, density estimation amounts to an optimization over a model defined by parameterized densities  $p(x|\theta)$. In both cases, optimization targets a real-valued \emph{loss function}, which is used to score the model $\theta$ when tested on a data point $z$. 

\subsection{Frequentist Learning}
The goal of frequentist learning consists in finding the model parameter vector $\theta$ that minimizes the \emph{training loss} evaluated on the data set $\mathcal{D}$, i.e.,
\begin{align}
    \hat{\mathcal{L}}(\theta,\mathcal{D})=\sum_{z\in\mathcal{D}}\ell(\theta,z). \label{emp_risk}
\end{align}
This optimization follows the \emph{empirical risk minimization} (ERM) principle. Accordingly, the frequentist solution is a \emph{single} model parameter $\theta^\text{freq} \in\Theta$ that minimizes the training loss, i.e.,
\begin{align}
    \theta^\text{freq} = \argmin_{\theta\in\Theta}  \hat{\mathcal{L}}(\theta,\mathcal{D}).
    \label{eq:freq_learning}
\end{align} To simplify the discussion, we assume that the solution to the ERM problem is unique, although this does not affect the generality of the presentation.

ERM is motivated by the fact that the training loss (\ref{emp_risk}) is a finite-sample approximation of the true, unknown, \emph{population loss}
\begin{align}
  \mathcal{L}(\theta)=\mathbb{E}_{\nu(z)}[\ell(\theta,z)], \label{pop_risk}
\end{align}which averages the loss over the target, and here also sampling, distribution $\nu(z)$. The discrepancy between the population loss and its approximation given by the training loss introduces uncertainty regarding the optimal model parameter 
\begin{align}
    \theta^* = \argmin_{\theta\in\Theta} \mathcal{L}(\theta), 
    \label{true_minimizer}
\end{align}which is also assumed to be unique to simplify the discussion. 
The error between the optimal solution $\theta^*$  and the frequentist solution $\theta^\text{freq}$ is a form of \emph{epistemic uncertainty}, which can be reduced by increasing the size of the data set $\mathcal{D}$. 

In practice, the short stationarity intervals of the data-generating distributions associated with wireless communications often limit the size of training data sets. In this scarce data regime, epistemic uncertainty may be significant. By selecting a single model, frequentist learning neglects epistemic uncertainty as it discards information about other plausible models that fit training data almost as well as the ERM solution (\ref{eq:freq_learning}). As a result, frequentist learning is known to lead to poorly calibrated decision \cite{guo2017calibration,cohen2021learning}, resulting in over- or under-confident outputs that may cause important reliability issues. 

\subsection{Bayesian Learning}
Bayesian learning  adopts a probabilistic reasoning framework by scoring all members in the model class by means of a distribution $q(\theta)$ over the model parameter space $\Theta$. Through this distribution, Bayesian learning summarizes information obtained from data $\mathcal{D}$, as well as prior knowledge about the problem, e.g., about the scale of the optimal model parameter vector $\theta^*$ or about sparsity patterns in $\theta^*$. 

Mathematically, given a prior distribution $p(\theta)$ on the model parameter space, Bayesian learning can be formulated as the minimization of the \emph{free energy criterion} 
\begin{align}
    \hat{\mathcal{J}}(q) = \mathbb{E}_{ q(\theta)}[\hat{\mathcal{L}}(\theta,\mathcal{D})] +\frac{1}{\beta}\text{KL}(q(\theta)||p(\theta)),
    \label{eq:free_energy}
\end{align}
where $\text{KL}(q(\theta)||p(\theta))$ denotes the Kullback–Leibler (KL) divergence between the posterior distribution $q(\theta)$ and a \emph{prior} distribution $p(\theta)$, i.e. 
\begin{align}
\text{KL}(q(\theta)||p(\theta))=\mathbb{E}_{q(\theta)}\left[\log\left(\frac{q(\theta)}{p(\theta)}\right)\right],
\end{align} while $\beta>0$ is a constant, also known as inverse temperature. Accordingly, through problem 
\begin{align}
    \minimize_{q}\hat{\mathcal{J}}(q),
    \label{eq:free_energy_min}
\end{align}
Bayesian learning minimizes a weighted sum of the average training loss and of the discrepancy with respect to the prior distribution $p(\theta)$.

The KL term in the free energy (\ref{eq:free_energy}) plays an essential role in differentiating between Bayesian learning and frequentist learning  for small data set sizes. In fact, the KL divergence term acts as a regularizer, whose influence on the solution of problem (\ref{eq:free_energy_min}) is inversely proportional to the data set size $n$. When the regularizer is removed, i.e., when we set $\beta \rightarrow \infty$,  the solution of the problem (\ref{eq:free_energy_min}) reduces to the frequentist solution (\ref{eq:freq_learning}). More precisely, the distribution $q(\theta)$ that solves problem (\ref{eq:free_energy_min}) reduces to a point distribution concentrated at $\theta^\text{freq}$.

The optimization (\ref{eq:free_energy_min}) of the free energy criterion (\ref{eq:free_energy}) can be theoretically justified through the \emph{PAC Bayes} generalization framework. In it, the KL term is proved to quantify an upper bound on the  discrepancy between training loss  and population loss on average with respect to the random draws of the model parameter vector $\theta \sim q(\theta)$. Mathematically, the free energy provides an upper bound on the average population loss (when neglecting constants that are inessential for optimization), i.e.,
\begin{align}
    \mathbb{E}_{q(\theta)}\left[\mathcal{L}(\theta)\right]\leq \hat{\mathcal{J}}(q)+ \textrm{const.} \label{free_bound}
\end{align}
As we have discussed in the previous subsection, epistemic uncertainty is caused by the difference between training and population losses, and hence between the corresponding minimizers (\ref{true_minimizer}) and (\ref{eq:freq_learning}). By incorporating a bound on this error, the free energy  criterion (\ref{eq:free_energy}) unlike the frequentist training loss  (\ref{emp_risk}), provides a way to account for epistemic uncertainty.

Specializing the problem (\ref{eq:free_energy_min})  to the \emph{log-loss} \begin{equation} \ell(x,y,\theta)=-\log p(y|x,\theta) \label{log_loss_gibbs} \end{equation} for supervised learning, and \begin{equation} \ell(x,\theta)=-\log p(x|\theta) \label{log_loss_unsup}\end{equation} for density estimation, the minimization of the free energy in \eqref{eq:free_energy_min} leads to the $\beta$-\emph{tempered posterior distribution}
\begin{align}
    q^{Bayes}(\theta|\mathcal{D})\propto \prod_{(x,y)\in\mathcal{D}} p(\theta) p(y|x,\theta)^\beta \label{bayes_pos}
\end{align}for supervised learning, and a similar expression applies to unsupervised learning for density estimation. The distribution (\ref{bayes_pos}) reduces to the standard posterior distribution when $\beta=1$. In practice, computing the posterior distribution, or more generally solving problem (\ref{eq:free_energy_min}), are computationally prohibitive tasks. A common approach  to address this issue is through \emph{variational inference (VI)} \cite{blei2017variational}. VI limits the scope of the optimization over a tractable set of distributions $q(\theta)$, such as jointly Gaussian variables with free mean and covariance parameters. 

Let us now assume that we have obtained a distribution $q(\theta)$ as a, generally approximate, solution of problem (\ref{eq:free_energy_min}). We focus first on supervised learning. Given a test input $x$, the \emph{ensemble predictor} obtained from distribution $q(\theta)$ is given by \begin{align}
    p(y|x,q)=\mathbb{E}_{ q(\theta)}[p(y|x,\theta)]. \label{ens_predictor}
\end{align}The average in (\ref{ens_predictor}) is in practice approximated by drawing multiple, say $m$, samples $\theta \sim q(\theta)$ from distribution $q(\theta)$, obtaining the \emph{$m$-sample predictor}\begin{align}
    p(y|x,\theta_1,...,\theta_m)=\frac{1}{m} \sum_{i=1}^{m} p(y|x,\theta_i),\label{ens_m_predictor}
\end{align}where samples $\theta_i$ are generated i.i.d. from distribution $q(\theta)$ for $i=1,...,m$, which we write as $\theta_1,...,\theta_m \sim q(\theta)^{\otimes m}$.

In the case of density estimation, the ensemble density $p(x|q)$ is similarly defined as \begin{align}
    p(x|q)=\mathbb{E}_{ q(\theta)}[p(x|\theta)], \label{ens_predictor1}
\end{align} 
which can be approximated as
\begin{align}
    p(x|\theta_1,\mydots,\theta_m)=\frac{1}{m} \sum_{i=1}^{m} p(x|\theta_i), \label{ens_predictor_m_unsup}
\end{align} 
with $\theta_1,\mydots,\theta_m\sim q(\theta)^{\otimes m}$.
Henceforth, when detailing expressions for supervised learning, it will be implied that the corresponding formulas for density estimation apply by replacing $p(y|x,\theta)$ with $p(x|\theta)$ as done above to define ensemble predictors.




Given a distribution $q(\theta)$, we define the \emph{$m$-sample log-loss} as
\begin{align}
    \ell(x,y,\theta_1,...,\theta_m)=-\log(p(y|x,\theta_1,...,\theta_m)), \label{ens_risk}
\end{align}which measures the log-loss of the $m$-sample predictor (\ref{ens_m_predictor}).

\textit{Example 1:} To illustrate the difference between the frequentist and Bayesian learning paradigms, let us consider the  problem of estimating the probability distribution of the channel gain of a scalar wireless channel. This is an example of unsupervised learning for density estimation. Let us assume that the channel gain density follows a true, unknown, target distribution given by the mixture of two Gaussians $\nu(x)= 0.7\mathcal{N}(x|0.5,0.05)+0.3\mathcal{N}(x|0.8,0.02)$. 
This is shown in the top part of Fig.  \ref{fig:toy_example} as a dashed green line. The two components may correspond to line-of-sight (LOS)  and non-line-of-sight (NLOS) propagation conditions \cite{xiao2013identification}. We fix a Gaussian model class $p(x|\theta)=\mathcal{N}(x|\theta,0.25)$ and a prior distribution $p(\theta)=\mathcal{N}(\theta|-5,5)$. Given the data points represented as crosses in the top part of Figure  \ref{fig:toy_example}, the estimated distribution obtained by frequentist learning is reported as a dash-dotted black curve in the top panel. In contrast, Bayesian learning returns the posterior distribution (\ref{bayes_pos}), which in turn yields the ensemble density (\ref{ens_predictor}). The distributions are shown in the top and bottom parts of the Figure \ref{fig:toy_example}, respectively for inverse temperature parameters $\beta=\{1,0.1\}$. The Bayesian predictive distribution is still unimodal but it has a larger variance, which results from the combination of multiple Gaussian models according to the Bayesian posterior that does not collapse to a point distribution in virtue of the KL regularization term whose influence is controlled by $\beta$. 
\hfill $\blacksquare$

\section{Robust Bayesian Learning}

\label{sec:RobBayes}
As we have seen in the previous section, Bayesian learning optimizes the free energy by tackling problem (\ref{eq:free_energy_min}). By (\ref{free_bound}), the free energy provides a bound on the population loss  as a function of the training loss when averaging over the distribution $q(\theta)$ in the model parameter space \cite{catoni2007pac}. This approach has two important limitations: \begin{itemize}
\item \emph{Model misspecification}: The bound (\ref{free_bound}) provided by the free energy is known to be loose in the presence of model misspecification. Model misspecification occurs when the assumed probabilistic model $p(y|x,\theta)$ cannot express the conditional target distribution $\nu(y|x)=\nu(x,y)/\nu(x)$, where $\nu(x)=\int\nu(x,y)dy$ \cite{masegosa2020learning,morningstar2020pac}. This causes the $\beta$-tempered posterior distribution to be generally suboptimal when the model is misspecified \cite{knoblauch2019generalized}.
\item \emph{Discrepancy between sampling and target distributions}: The sampling distribution $\nu_s(z)$ that underlies the generation of the training data set $\mathcal{D}$ may not match the target distribution $\nu(x)$ used to test the trained model due to the presence of outliers in the training data. This discrepancy is not accounted for in the derivation of the free energy criterion, causing Bayesian learning to be suboptimal in the presence of outliers \cite{knoblauch2019generalized}. 
\end{itemize} We observe that the two causes of suboptimality outlined in the previous paragraph are distinct. In fact, model misspecification may reflect the ignorance of the learner concerning the data generation process, or it may be caused by constraint on the computational resources of the device implementing the model. In contrast, the presence of outliers amounts to an inherent source of distortion in the data, which cannot be removed even if the learner acquired more information about the data generation process or more computing power. In this section, we review robust Bayesian learning solutions that address these two issues.

\subsection{$(m,1)$-Robust Bayesian Learning Against Model Misspecification}
\label{sec:mis}

In this subsection, we describe a recently proposed method that robustifies Bayesian learning against model misspecification. We start by providing a formal definition of misspecification. Recall that we are focusing on supervised learning, but the presentation also applies to density estimation by replacing the discriminative model $p(x|y,\theta)$ with the density model $p(x|\theta)$.
\begin{definition}[Misspecification]
\label{ass:mis}
A model class $\mathcal{F}=\{p(y|x,\theta):\theta\in\Theta\}$ is said to be misspecified with respect to the target distribution $\nu(x,y)$ whenever there is no model parameter vector $\theta\in \Theta$ such that $\nu(y|x)=p(y|x,\theta)$, where $\nu(y|x)$ is the conditional target distribution obtained from the joint target distribution $\nu(x,y)$.
\end{definition} Under model misspecification, the free energy criterion has been shown to yield a loose bound (\ref{free_bound}) on the population loss obtained by the ensemble predictor (\ref{ens_predictor}) \cite{morningstar2020pac}. 

To address this problem, the $m$-sample free energy criterion was introduced in \cite{morningstar2020pac}, whose minimization yields $(m,1)$-robust learning. The reason for the notation `` $(m,1)$" will be made clear in the next two subsections. The key observation underlying this approach is that the training loss  $\hat{\mathcal{L}}(\theta,\mathcal{D})$ in the standard free energy  (\ref{eq:free_energy}) does not properly account for the performance of ensemble predictors. In fact, the log-loss of an $m$-sample ensemble predictor is given by $\ell(x,y,\theta_1,...,\theta_m)$ in (\ref{ens_risk}), and not by the log-loss $\ell(x,y,\theta)$ in (\ref{log_loss_gibbs}). Accordingly, the $m$-sample free energy is obtained by replacing the training loss $\hat{\mathcal{L}}(\theta,\mathcal{D})$ in the free energy 
    (\ref{eq:free_energy}) with the \emph{$m$-sample training loss} 
\begin{align}
    \label{eq:multi_log_loss}
    \hat{\mathcal{L}}(\theta_1,\ldots,\theta_m, \mathcal{D}) & =\sum_{(x,y) \in \mathcal{D}} \ell(x,y,\theta_1,...,\theta_m) \nonumber \\ & = -\sum_{(x,y) \in \mathcal{D}} \log \left(\sum^m_{i=1}\frac{p(y|x,\theta_i)}{m}\right).
\end{align} Furthermore, the \emph{$m$-sample free energy} is defined as
\begin{align}
    \hat{\mathcal{J}}^m(q)&= \mathbb{E}_{q(\theta)^{\otimes m}}[\hat{\mathcal{L}}(\theta_1,\ldots,\theta_m, \mathcal{D})]+ \frac{m}{\beta}\text{KL}(q(\theta)||p(\theta)),
    \label{eq:pac_m_bayes}
\end{align} in which the $m$-sample training loss is averaged over the distribution of the $m$ samples $\theta_1,...,\theta_m\sim q(\theta)^{\otimes m}$ used in the ensemble predictor (\ref{ens_m_predictor}). We note that the $m$-sample free energy coincides with the standard free energy (\ref{eq:free_energy}) for $m=1$.

Finally, the \emph{$(m,1)$-robust Bayesian learning} problem is defined by the optimization 
\begin{align}
    \minimize_{q}\hat{\mathcal{J}}^m(q).
    \label{eq:M_free_energy_min}
\end{align}

\textit{Example 1 (continued):} Let us return to Example 1. The problem  is characterized by model misspecification since the target distribution $\nu(x)$ is a mixture of two Gaussian components, while the model class comprises only unimodal Gaussian models $p(x|\theta)$. In contrast to standard Bayesian learning, the ensemble density (\ref{ens_m_predictor}) obtained with the distribution $q(\theta)$ returned by $(m,1)$-robust Bayesian learning for $m=10$ (red curve in the top panel) is able to take advantage of ensembling to approximate both the NLOS and LOS components of the target distribution.  \hfill $\blacksquare$

\subsection{$(1,t)$-Robust Bayesian Learning Against Outliers}
\begin{figure}
    \centering
    \hspace{-1em}
    \includegraphics[width=0.5\textwidth]{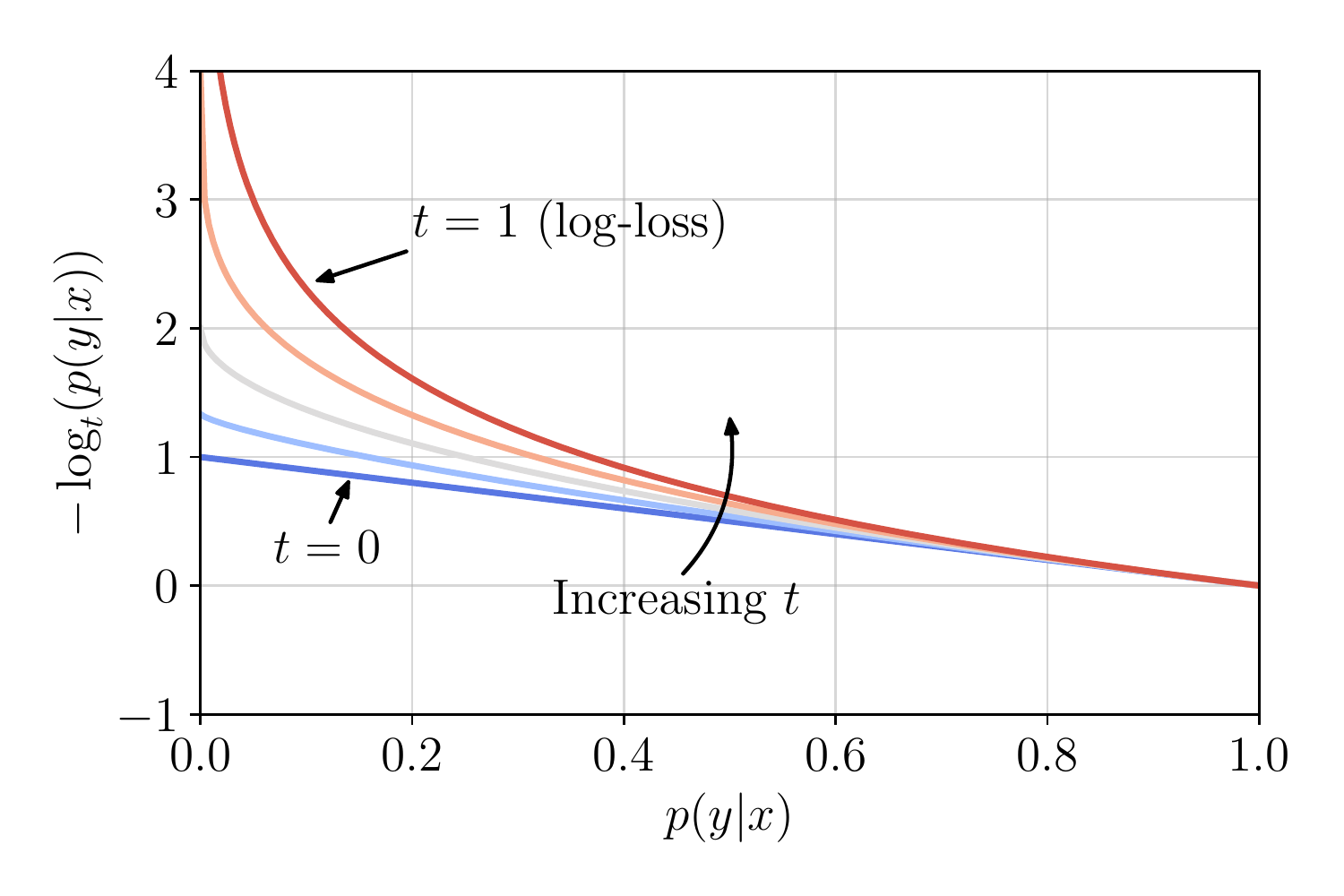}
    \caption{\blue{$t$-log-loss $-\log_t(p(y|x)$ as a function of the predictive probability $p(y|x)$ for different values of $t$. For $t=1$, the $t$-log-loss coincides with the conventional log-loss.  A sample $(x,y)$ with a low predictive probability $p(y|x)\to0$ is assigned an unbounded $\log$-loss value. In contrast, for $t<1$, the $t$-log-loss is bounded by $(1-t)^{-1}$, limiting the influence of outliers.}}
    \label{fig:t_log_fig}
\end{figure}
\label{sec:rob_div}


We now turn to methods that robustify Bayesian learning against the presence of outliers in the training set. As in \cite{Huber}, we model the presence of outliers by assuming that the training data is generated from a sampling distribution $\nu_s(x,y)$ that is given by the contamination of the target distribution $\nu(x,y)$ by an \emph{out-of-distribution (OOD) distribution} $\xi(x,y)$. A formal definition follows. 
\begin{assumption}[Outliers]
\label{ass:out}
The sampling distribution is given by 
\begin{align}
    \nu_s(x,y)=(1-\epsilon)\nu(x,y)+\epsilon\xi(x,y)
    \label{cont_mod}
\end{align}
where $\nu(x,y)$ is the target distribution; $\xi(x,y)$ is the OOD distribution accounting for the presence of outliers; and $\epsilon\in[0,1]$ denotes the contamination ratio.
\end{assumption} 
In order for model (\ref{cont_mod}) to be meaningful, one typically assumes that the OOD measure $\xi(x,y)$ is large for pairs of $(x,y)$ at which the target measure $\nu(x,y)$ is small. This ensures that outlying data points $(x,y) \sim \xi(x,y)$ tend to be in part of the domain that is not covered by the target distribution.

The performance of both frequentist and Bayesian learning is known to be sensitive to outliers when the log-loss is adopted to evaluate the training loss. This sensitivity is caused by the unbounded value of the log-loss (\ref{log_loss_gibbs}) when evaluated on anomalous data points to which the model assigns low probabilities $p(y|x,\theta)$. This is illustrated in Figure \ref{fig:t_log_fig} for a general conditional distribution $p(y|x)$.  A number of  papers have proposed to mitigate the effect of outliers by replacing the log-loss in favor of more robust losses  \cite{basu1998robust,ghosh2016robust,fujisawa2008robust,nakagawa2020robust,jewson2018principles}. 

A well-explored solution is to adopt the $t$-log-loss. For for a model $p(y|x,\theta)$, the \emph{$t$-log-loss} is defined as 
\begin{equation}
    -\log_t(p(y|x,\theta)):=-\frac{1}{1-t}\left(p(y|x,\theta)^{1-t}-1\right) \, \text{ for } p>0,
    \label{t_log_def}
\end{equation}
where $t\in[0,1)\cup(1,\infty)$; and 
\begin{equation}
   -\log_1(p(y|x,\theta)):=-\log(p(y|x,\theta)) \, \text{ for } p>0.
    \label{log_def}
\end{equation} 
By (\ref{log_def}) the standard log-loss is obtained with $t=1$, while for $t<1$ the associated loss function is bounded by $(1-t)^{-1}$, as shown in Figure \ref{fig:t_log_fig}.

Using the $t$-log-loss in lieu of the standard log-loss in the training loss (\ref{emp_risk}) we obtain the $t$-training loss 
\begin{align}
    \hat{\mathcal{L}}_t(\theta, \mathcal{D}) = - \sum_{(x,y) \in \mathcal{D}} \log_t \left(p(y|x,\theta)\right), 
\end{align}
which leads to the corresponding \emph{$t$-free energy} 
\begin{align}
        \hat{\mathcal{J}}_t(q)&=\mathbb{E}_{q(\theta)}[\hat{\mathcal{L}}_t(\theta, \mathcal{D})]+ \frac{1}{\beta}\text{KL}(q(\theta)||p(\theta)).
    \label{eq:robust_t_bayes}
\end{align} 
Accordingly, \emph{$(1,t)$-robust Bayesian learning} is defined by the minimization \cite{morningstar2020pac}
\begin{align}
    \minimize_{q}\hat{\mathcal{J}}_t(q).
    \label{eq:robus_free_energy_min}
\end{align}

\begin{figure}
   \includegraphics[width=1\linewidth]{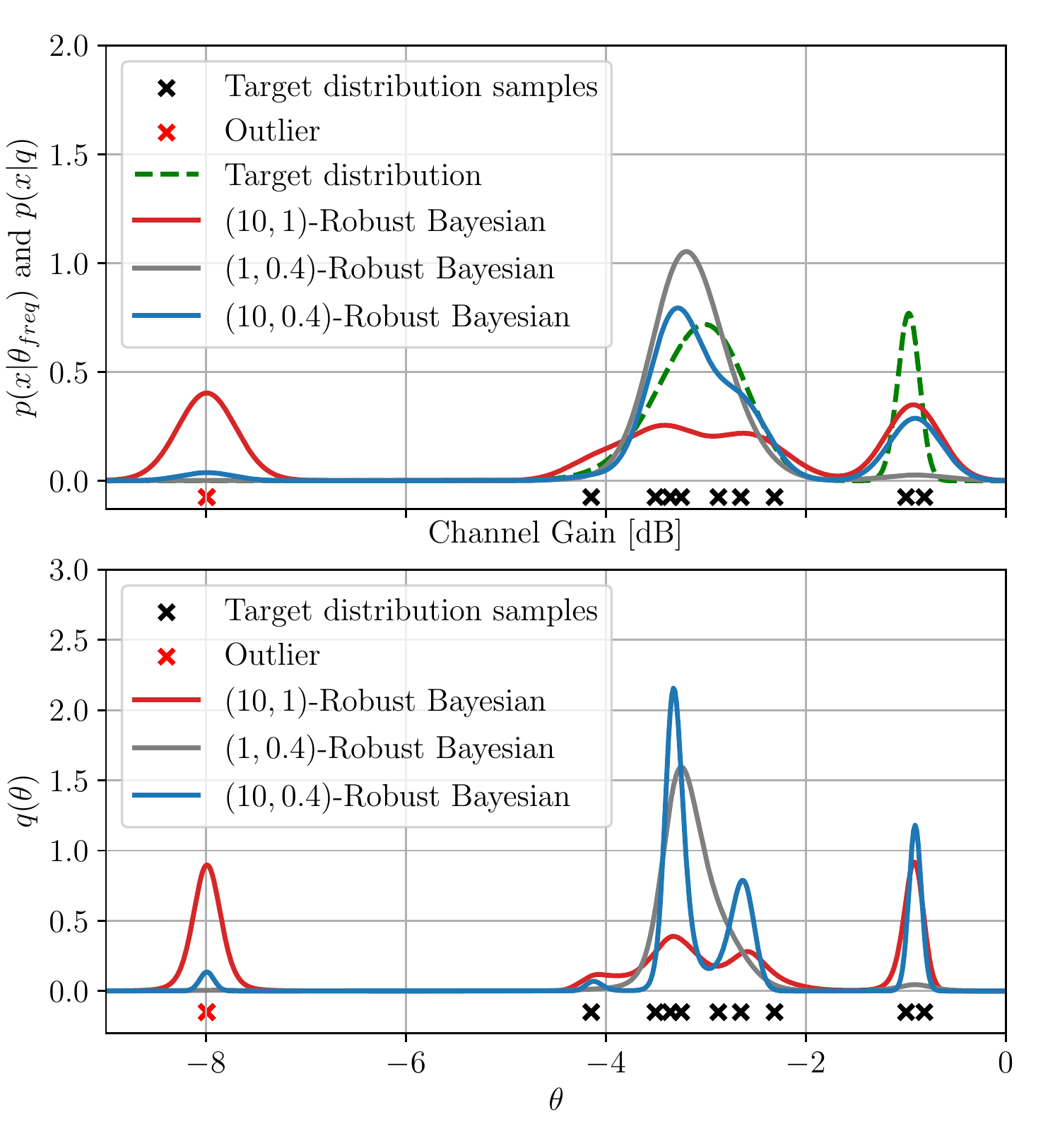}
    \caption{\blue{Estimated distribution over channel gains (top panel) and posterior distribution over the model parameter $\theta$ (bottom panel) of a density model trained following $(m,1)$-robust Bayesian learning, the $(1,t)$-robust Bayesian learning and the $(m,t)$-robust Bayesian learning. The training data set, represented  as crosses, comprises samples from the sampling distribution $\nu(x)$ (black) and an outlier (red).}} 
    \label{fig:toy_example_2}
\end{figure}
\blue{
\textit{Example 2:}  To highlight the effect of outliers, we consider the same channel gain estimation problem described in Example 1, but we now assume that the original training data set (black crosses) is contaminated by an outlying data point (red cross). The $(m,1)$-robust Bayesian learning solution (red curve with $m=10$) is based on the standard log-loss and is observed to be significantly affected by the presence of the outliers. As a result, the estimated distribution for the $(m,1)$-robust Bayesian learning concentrates a relevant fraction of its mass around the outlier. In contrast, the $(1,t)$-robust Bayesian solution (gray curve) with $t=0.4$ is less influenced by the outlying data point. However, like Bayesian learning, it is not able to take advantage of ensembling and to approximate both LOS and NLOS components. This observation justifies the $(m,t)$-robust Bayesian learning approach described next. \hfill $\blacksquare$}

\subsection{$(m,t)$-Robust Bayesian Learning Against Model Misspecification and Outliers}
\label{sec:robpacm}
 To concurrently address model misspecification and the presence of outliers, reference \cite{zecchin2022robust} formally introduced $(m,t)$-robust Bayesian learning, which minimizes a free energy metric integrating both $m$-sample predictors and the $t$-log-loss. To describe it, let us first define the \emph{$(m,t)$-training loss}\begin{align}
    \hat{\mathcal{L}}_t(\theta_1,\ldots,\theta_m, \mathcal{D}) = - \sum_{(x,y) \in \mathcal{D}} \log_t \left(\sum^m_{i=1}\frac{p(y|x,\theta_i)}{m}\right), 
\end{align}which is obtained from the $m$-sample training loss (\ref{eq:multi_log_loss}) by replacing the log-loss with the $t$-log-loss. The \emph{$(m,t)$-free energy} is accordingly defined as\begin{align}
        \hat{\mathcal{J}}^m_t(q)&=\mathbb{E}_{q(\theta)^{\otimes m}}[\hat{\mathcal{L}}_t(\theta_1,\ldots,\theta_m, \mathcal{D})]+ \frac{m}{\beta}\text{KL}(q(\theta)||p(\theta)), 
    \label{eq:pac_m_t_bayes}
\end{align} and \emph{$(m,t)$-robust Bayesian learning} amounts to the minimization 
\begin{align}
    \minimize_{q}\hat{\mathcal{J}}_t^m(q).
    \label{eq:robust_M_free_energy_min}
\end{align}
Note that $(m,t)$-robust Bayesian learning recovers standard Bayesian learning by setting $t=1$ and $m=1$, as well as $(m,1)$-robust Bayesian learning with  $t=1$ and the $(1,t)$-robust Bayesian learning for $m=1$.

\blue{
\textit{Example 2 (continued):} Returning to Example 2, we now consider the performance of $(m,t)$-robust Bayesian learning for $m=10$ and $t=0.4$. The resulting distribution (blue line) with  $m=10$ and $t=0.4$ seems to be able to better to approximate the target distribution by reducing the effect of the outliers, while also taking advantage of ensembling to combat misspecification.}


\section{Robust and Calibrated Automatic Modulation Classification}
\label{exp:amc}
As a first application of robust Bayesian learning we consider the AMC problem. This is the task of classifying received baseband signals in terms of the modulation scheme underlying their generation. The relation between the received signal and the chosen modulation scheme is often mediated by complex propagation phenomena, as well as hardware non-idealities at both the receiver and the transmitter side. As a result,  model-based AMC methods often  turn out to be inaccurate because of the overly simplistic nature of the assumed models \cite{o2018over}. In contrast, machine learning based AMC has been shown to be extremely effective in correctly classifying received signals based on signal features autonomously extracted from data \cite{o2016convolutional}. We refer to \cite{zhou2020deep} and references therein for a comprehensive overview.

All prior works on learning-based AMC, reviewed in \cite{zhou2020deep}, have adopted frequentist learning. In this section, we consider the practical setting in which AMC must be implemented on resource-constrained devices,  entailing the use of small, and hence mismatched, models; and  the training data sets are characterized by the presence of outliers due to interference.

\subsection{Problem Definition and Performance Metrics}
The AMC problem can be framed as an instance of supervised classification, with the training data set $\mathcal{D}$ comprising pairs $(x,y)$ of discrete-time received baseband signal $x$ and modulation label $y$, with $\mathcal{Y}$ being the set of possible modulation schemes. Each training data point $(x,y)\in\mathcal{D}$ is obtained by transmitting a signal with a known modulation $y\in \mathcal{Y}$ over the wireless channel, and then recording the received discrete-time vector $x$ at the receiver end. The outlined procedure determines the unknown sampling distribution $\nu_s(x,y)$.



We evaluate the performance of AMC on a testing data set $\mathcal{D}_{te}$ in terms of accuracy and \emph{calibration}. To  describe calibration performance metrics, let us consider a predictive distribution $p(y|x)$, which may be the frequentist distribution $p(y|x,\theta^{\text{freq}})$, or the ensemble distribution (\ref{ens_m_predictor}) in the cases of Bayesian learning and robust Bayesian learning. A hard prediction $\hat{y}$ is obtained as the maximum-probability solution
\begin{equation}
\hat{y}=\argmax_{y\in\mathcal{Y}}p(y|x).
\end{equation} The corresponding \emph{confidence score} assigned by the predictor $p(y|x)$ is the probability $p(\hat{y}|x)\in[0,1]$. The calibration of a classifier measures the degree to which the confidence score $p(\hat{y}|x)\in[0,1]$ reflects the true probability of correct classification $P[\hat{y}=y|x]$ conditioned on the input $x$.  

We adopt the standard reliability diagrams \cite{degroot1983comparison} and the expected calibration error as diagnostic tools for the calibration performance \cite{guo2017calibration}. Both metrics require binning the output of the classifier confidence score $p(\hat{y}|x)$ into $M$ intervals of equal size, and then grouping the testing data points $(x,y)\in\mathcal{D}_{te}$ based on the index of the bin for the confidence score $p(\hat{y}|x)$. For each bin $\mathcal{B}_m$, the \emph{within-bin accuracy} is defined as
\begin{equation}
    \textrm{Acc}(\mathcal{B}_m)=\frac{1}{|\mathcal{B}_m|}\sum_{(x,y)\in\mathcal{B}_m}\mathbb{1}\{\hat{y}=y\},
\end{equation}which measures the fraction of test samples within the bin that are correctly classified; 
and the \emph{within-bin confidence} as
\begin{equation}
    \textrm{Conf}(\mathcal{B}_m)=\frac{1}{|\mathcal{B}_m|}\sum_{(x,y)\in\mathcal{B}_m}p(\hat{y}|x), 
\end{equation}which is the average confidence level for the test samples within the bin.

The \emph{reliability diagram} plots within-bin accuracy and within-bin confidence as a function of the bin index $m$. As a result, a reliability diagram visualizes the relation between confidence and accuracy of a predictor, establishing whether a classifier is over-confident  $\left(\textrm{Conf}(\mathcal{B}_m)>\textrm{Acc}(\mathcal{B}_m)\right)$, under-confident $\left(\textrm{Conf}(\mathcal{B}_m)<\textrm{Acc}(\mathcal{B}_m)\right)$ or well-calibrated  $\left(\textrm{Conf}(\mathcal{B}_m)\approx \textrm{Acc}(\mathcal{B}_m)\right)$. 

The \emph{expected calibration error (ECE)} summarizes the calibration performance of a classifier as a single number obtained as the weighted sum of the absolute difference between within-bin accuracy and within-bin confidence, namely
\begin{equation}
    \mathrm{ECE}=\sum^M_{m=1}\frac{|\mathcal{B}_m|}{\sum^M_{m=1}|\mathcal{B}_m|}\left|\textrm{Conf}(\mathcal{B}_m)- \textrm{Acc}(\mathcal{B}_m)\right|.
\end{equation}
By this definition, one can generally conclude that a lower ECE indicates a better calibrated predictor.


\begin{figure}
   \includegraphics[width=1\linewidth]{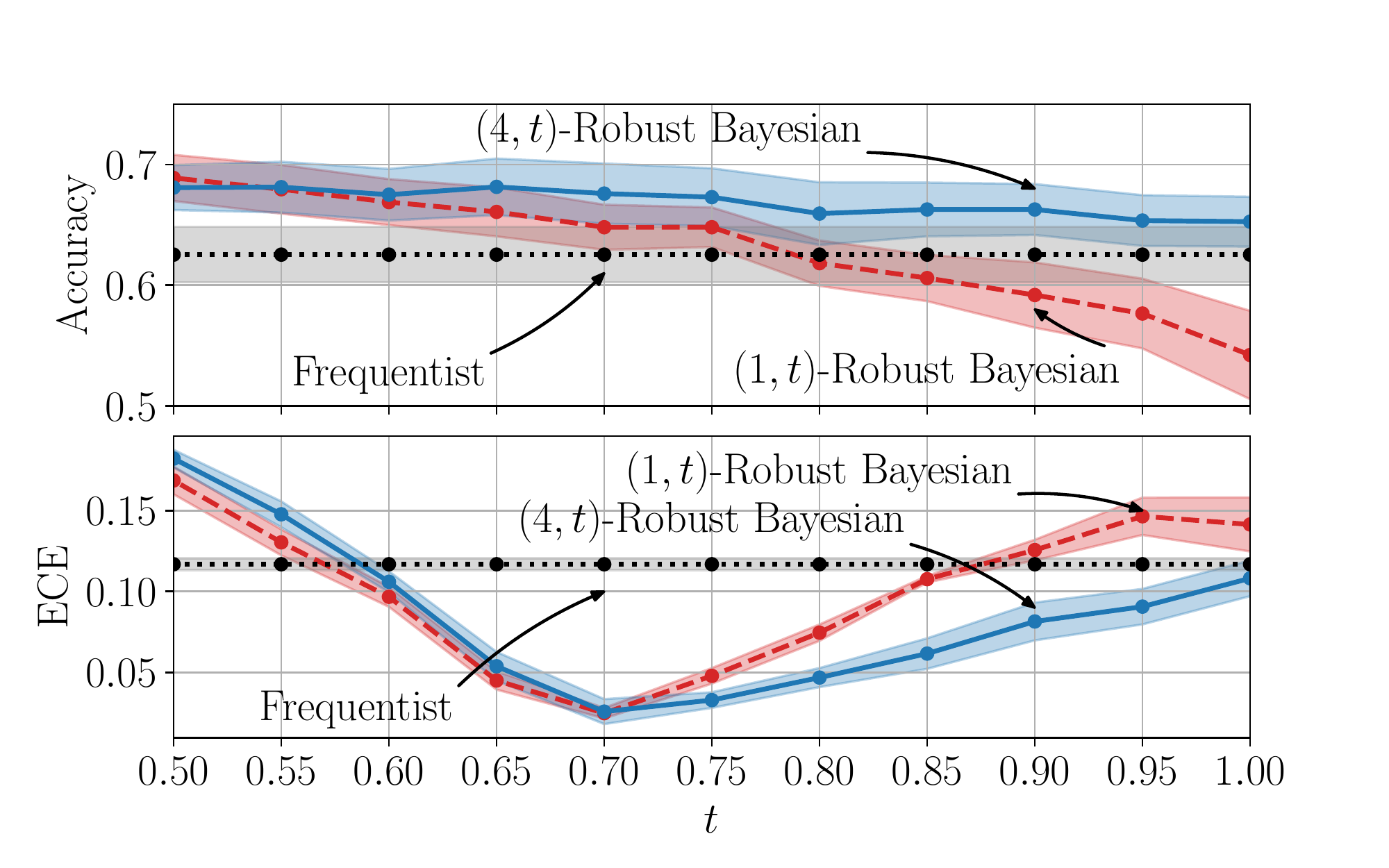}
   \caption{Average test accuracy and ECE for AMC over the DeepSIG: RadioML 2016.10A data set \cite{o2016convolutional} for frequentist and $(m,t)$-robust Bayesian learning as a function of the parameter $t$. The test set is free from interference, while the training set is subject to interference $(\epsilon=0.5)$. }
     \label{fig:ECE_ACC_AMS_plot} 
\end{figure}


\subsection{Data Set}

We adopt the \textit{DeepSIG: RadioML 2016.10A} data set \cite{o2016convolutional}. This is a synthetic data set that contains  220K vectors of I/Q samples of signals comprising 8 digital modulation schemes (BPSK, QPSK, 8PSK, 16QAM, 64QAM, BFSK, CPFSK) and 3 analog modulations (WB-FM, AM-SSB, AM-DSB). We focus on the problem of classifying the 8 digital modulation schemes using received signals recorded at different SNR levels ranging from $0$ dB to $18$ dB. Furthermore, we model the presence of \emph{interference} during training by generating an $\epsilon$-contaminated version of the original data set. In it, with probability  $\epsilon\in[0,1)$, the original training sample $x$ is summed to an interfering signal $x'$ picked uniformly at random from the data set. Note that the interfering signal can be possibly generated from a different modulation scheme.   Using Definition 2, the samples affected by interference represent \emph{outliers}, since no interference is assumed during testing. We consider $30\%$ of the available samples for training; $20\%$ of the samples for validation; and the remaining  $50\%$ for testing. The use of a small training data set is intentional, as we wish to focus on a regime characterized by data scarcity.

\subsection{Implementation}


We implement a lightweight convolutional neural network (CNN) architecture comprising of two convolutional layers  followed by two linear layers with 30 neurons each. The first convolutional layer has 16 filters of size $2\times3$, and the second layer has 4 filters of size $1\times2$. We adopt the Exponential Linear Unit (ELU) activation with parameter $\alpha=1$. The lightweight nature of the architecture is motivated by the strict computational and memory requirements at network edge devices. As a result, the CNN model is generally  \emph{misspecified}, in the sense that, following Definition 1, the complex relation between received signal and chosen modulation scheme cannot be exactly represented using the model.

In the training data set, half of the samples are affected by interference, i.e., $\epsilon=0.5$. For Bayesian learning, we adopt a Gaussian variational distribution $q(\theta)=\mathcal{N}(\theta|\mu,\Sigma)$ over the CNN model parameter vector  $\theta$. Accordingly, the mean $\mu$ and diagonal covariance matrix $\Sigma$ are optimized, while we fix the prior $p(\theta)=\mathcal{N}(\theta|0,I)$. Optimization for both frequentist and Bayesian methods is carried out via Adam with a learning rate $\eta=0.001$, and the reparametrization trick is implemented for Bayesian learning \cite{kingma2013auto}. In our experiments we set $\beta=0.01$. The number of samples used to evaluate the ensemble prediction (\ref{ens_m_predictor}) is $m=10$. Note that this may differ from the value of $m$ used to define the training criterion.
\subsection{Results}

In Figure \ref{fig:ECE_ACC_AMS_plot} we report the average test accuracy and ECE for frequentist and $(m,t)$-robust Bayesian with different values of $m$ as a function of $t$.  The main observation is that, with suitably chosen parameters $(m,t)$, robust Bayesian learning can outperform standard frequentist learning both in terms of accuracy and calibration for $t<1$. The smallest ECE is obtained by robust Bayesian learning for $t=0.7$, and it is five times smaller compared to the one obtained using conventional Bayesian learning ($t=1$). Overall,  $(m,t)$-robust Bayesian paradigm is able to improve the final accuracy by $5\%$ and to reduce the ECE by five times via  suitable choice of parameters $(m,t)$.


\begin{figure}
\begin{subfigure}{0.245\textwidth}
   \includegraphics[width=1\linewidth]{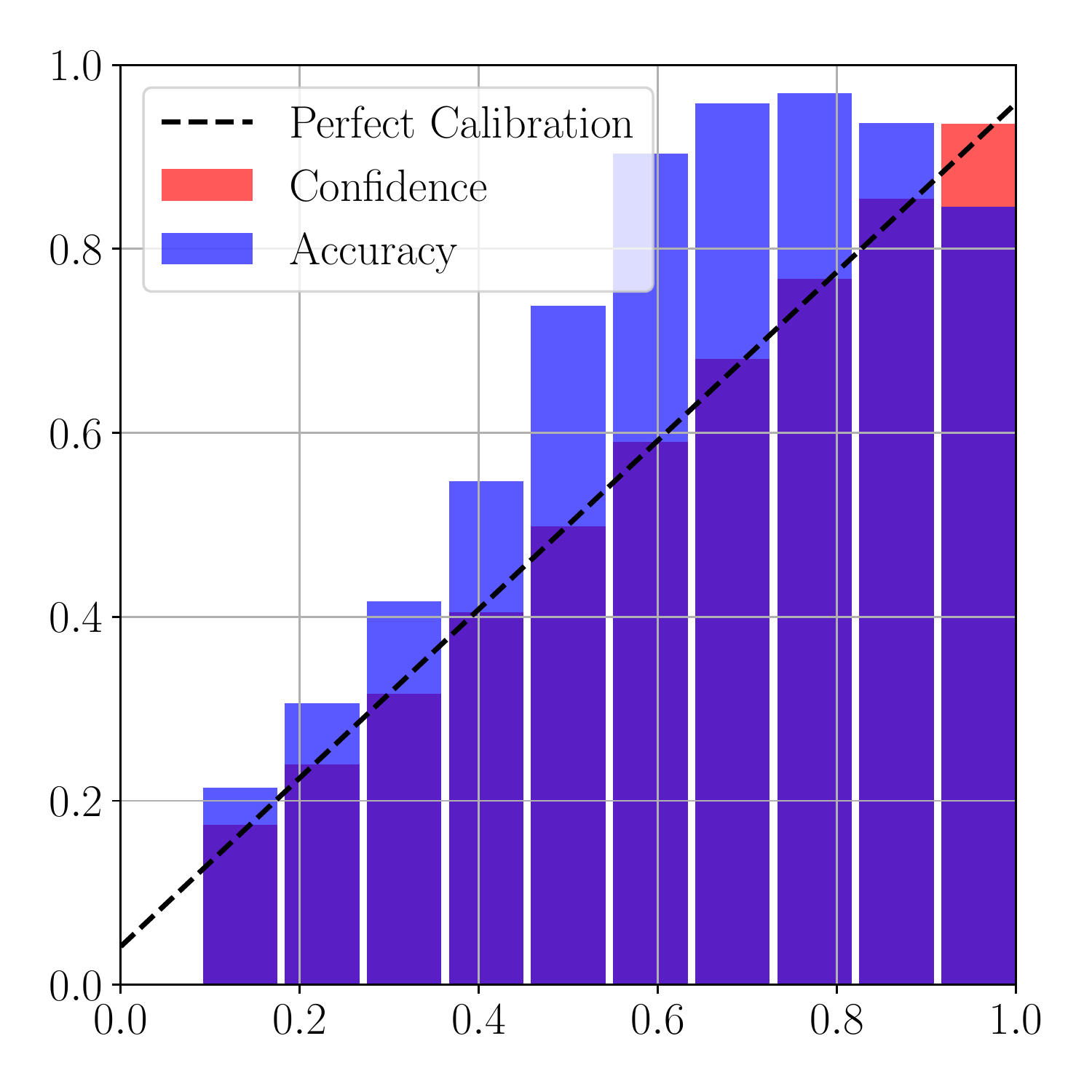}

    \caption{Frequentist Learning}
     \label{fig:Rel_plot_2} 
\end{subfigure}
\hspace{-1em}
\begin{subfigure}{0.245\textwidth}
   \includegraphics[width=1\linewidth]{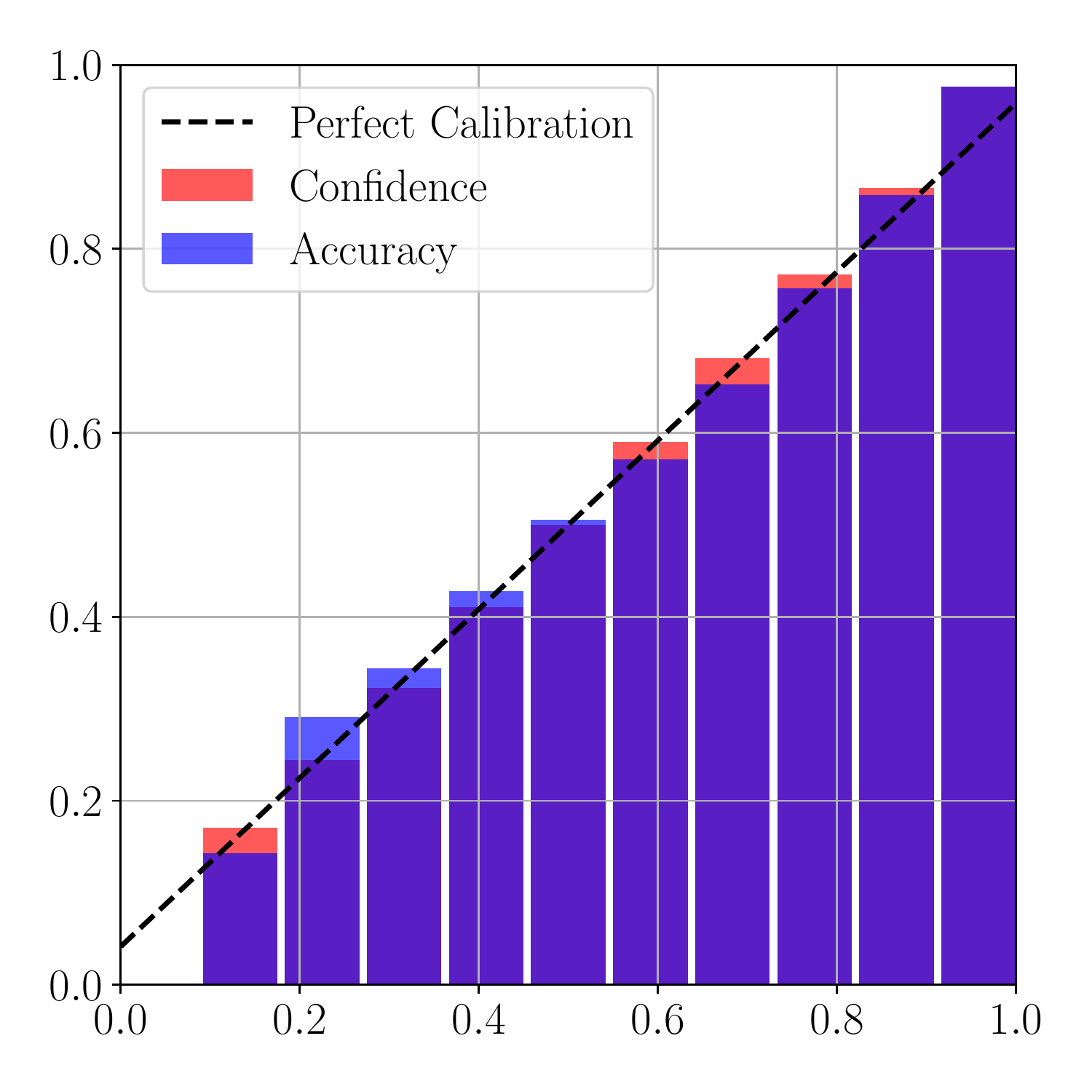}
    \caption{Robust Bayesian Learning}
    \label{fig:Rel_plot_3} 
\end{subfigure}
\caption{Reliability diagrams for frequentist (left) and $(m,t)$-robust Bayesian learning for $m=4$ and $t=0.7$ (right) for AMC over the DeepSIG: RadioML 2016.10A data set \cite{o2016convolutional}. }
\label{fig:Rel_curves}
\end{figure}


To further elaborate on the calibration performance, in Figure \ref{fig:Rel_curves} we compare the reliability diagrams obtained via frequentist and $(m,t)$-robust Bayesian learning for $m=4$ and $t=0.7$. While frequentist learning provides under-confident predictions, robust Bayesian learning offers well-calibrated predictions that consistently offer a small discrepancy between accuracy and confidence levels. 


\section{Robust and Calibrated RSSI-Based Localization}

\begin{table}[]
\centering
\caption{Test negative log-likelihood for RSSI localization (\ref{NNL_metric}) with $t=1$ and no outliers ($\epsilon=0$). The case $m=1$ corresponds to conventional Bayesian learning.}
\label{tab:KL}
\normalsize
\begin{tabular}{@{}lrrr@{}}
\toprule
                & \multicolumn{1}{c}{$m=1$} & \multicolumn{1}{c}{$m=2$} & \multicolumn{1}{c}{$m=10$} \\ \midrule
 \textit{SigfoxRural}    & $1.70\pm1.03$                         & $-0.43\pm0.61$                        & $\mathbf{-1.59\pm0.36}$                \\
 \textit{UTSIndoor}       & $4.33\pm2.32$                         & $2.25\pm1.69$                         & $\mathbf{2.17\pm1.76}$                    \\
 \textit{UJIIndoor}       & $4.86\pm1.02$                         & $2.74\pm0.46$                         & $\mathbf{1.44\pm0.33}$                    \\ \bottomrule
\end{tabular}

\end{table}
\begin{figure}
\begin{subfigure}{0.245\textwidth}
   \includegraphics[width=1\linewidth]{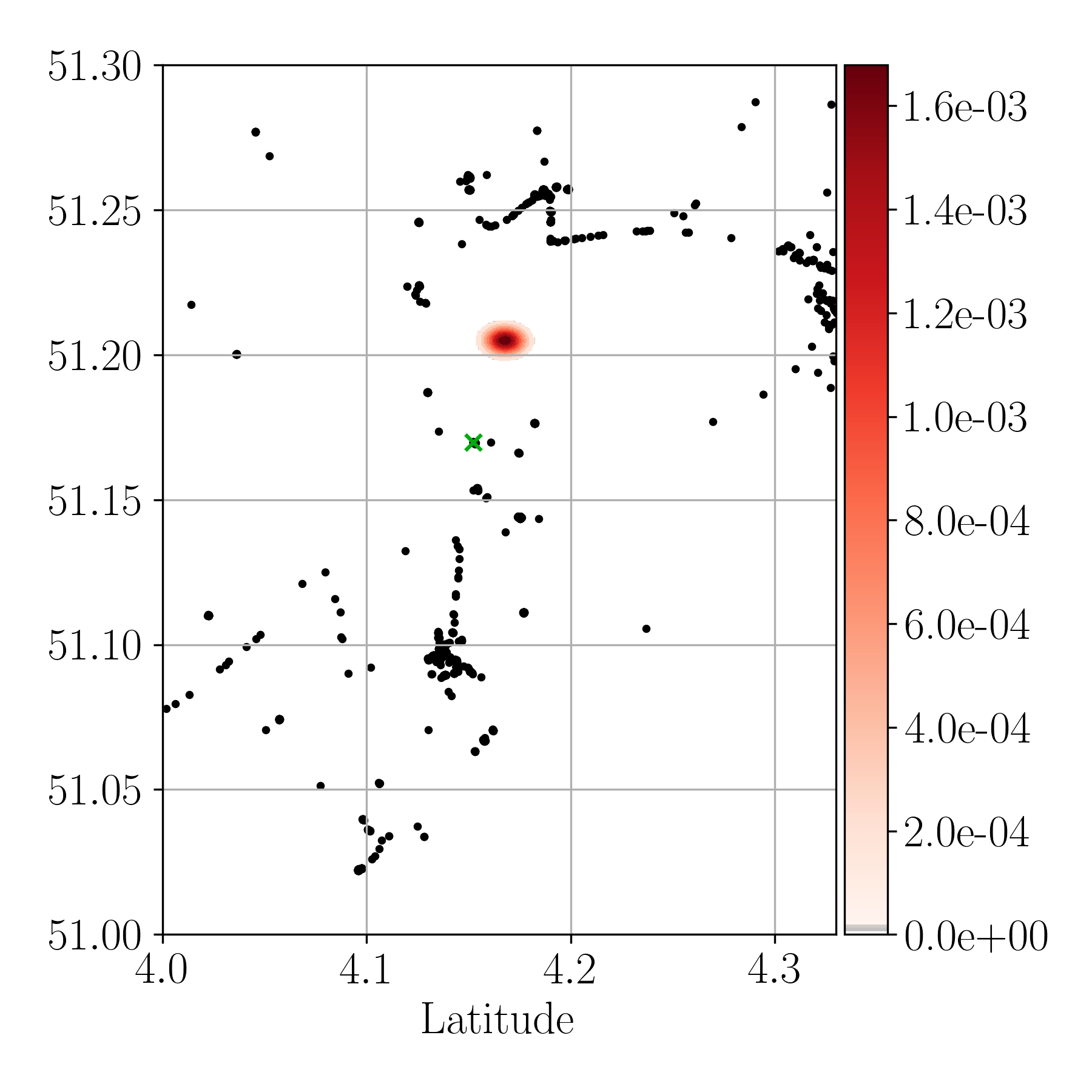}
   \caption{Bayesian Learning}
   \label{fig:Ng1} 
\end{subfigure}
\hspace{-1em}
\begin{subfigure}{0.245\textwidth}
   \includegraphics[width=1\linewidth]{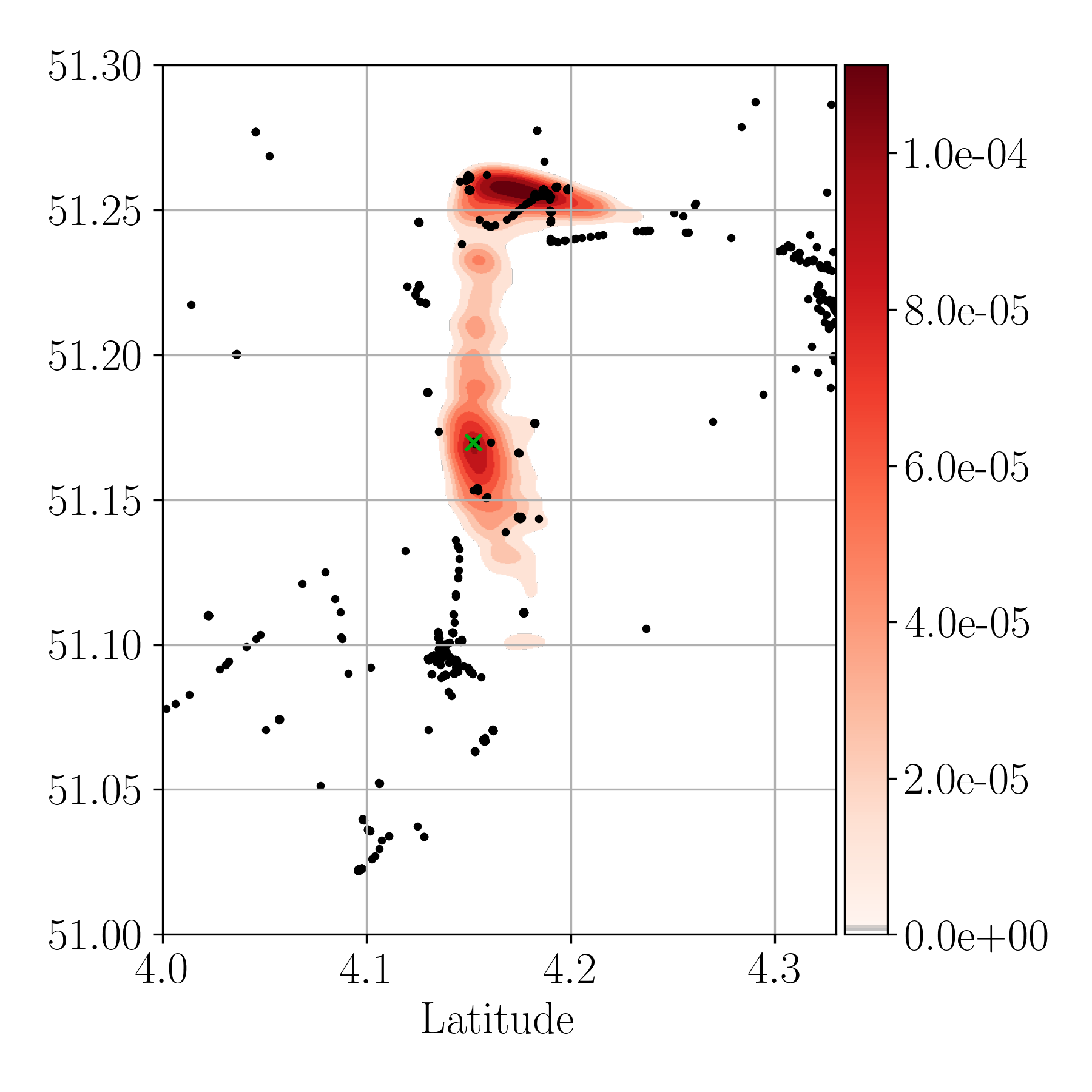}
    \caption{Robust Bayesian Learning}
   \label{fig:Ng2}
\end{subfigure}
\caption{Predictive distribution $p(y|x)$ as a function of the estimated position of the transmitter $y$, where $x$ is the RSSI vector associated to the true location shown as a green cross. The black dots correspond to the locations recorded in the  \textit{SigfoxRural} data set. The left panel shows the predictive distribution for Bayesian learning, while the right panel depicts the predictive distribution for  $(m,t)$-robust Bayesian learning  with $m=10$ and $t=1$. No outliers are considered in the training set, i.e., $\epsilon=0$.}\label{fig:RSSI}
\end{figure}

\label{exp:RSSI}

In this section, we turn to the problem of localization.  In outdoor environments, accurate localization information of a wireless device can be obtained leveraging the global navigation satellite system (GNSS). However, the performance of satellite-based positioning is severely degraded in indoor environments \cite{mautz2009overview}, and its power requirements are not compatible with IoT application characterized by ultra-low power consumption \cite{aernouts2018sigfox}. For this reason, alternative techniques have been investigated that rely on so-called \emph{channel fingerprints}, i.e., feature extracted from the received wireless signals \cite{pecoraro2018csi}.

Among such methods, the use of \emph{received signal strength indicators} (RSSI) measured at multiple wireless access points has been shown to provide an accessible, yet informative, vector of features. Owing to the complexity of defining explicit models relating the device location $y\in\mathcal{Y}$ with the RSSI-measurements vector $x\in\mathcal{X}$,  data-driven RSSI-based localization techniques have been recently explored  \cite{hoang2019recurrent,sinha2019comparison}. The outlined prior work in this area has focused on machine learning models trained using the conventional frequentist approach. 

In this section, we study a setting in which the training data set is collected using noisy, e.g., \emph{crowd-sourced}, fingerprints. As such, the training set contains outliers. Furthermore, we aim at developing strategies, based on robust Bayesian learning, which can offer accurate localization, while also properly quantifying residual uncertainty.

\subsection{Problem Definition and Performance Metrics}

The RSSI-based localization problem is a supervised regression task. In it, a training sample $(x,y)$ is obtained by measuring the RSSI fingerprint $y$ corresponding to the transmission of a reference signal at a device located at a known position $x$. The general goal is to train a machine learning model $p(y|x)$ to predict the location $y$ associated  to a RSSI vector $x$ so as to optimize accuracy and uncertainty quantification. 

Given a test data set $\mathcal{D}_{te}$ and assuming that the predictive location is the mean of the predictive distribution, i.e. $\Bar{y}=\mathbb{E}_{p(y|x)}[y]$, we adopt the \emph{mean squared error (MSE)} metric
\begin{equation}
    \text{MSE}(\mathcal{D}_{te},p)=\frac{1}{|\mathcal{D}_{te}|}\sum_{(x,y)\in\mathcal{D}_{te}}\norm{y-\bar{y}}_2
    \label{MSE_metric}
\end{equation} as a measure of accuracy. Furthermore, in order to estimate the residual uncertainty about $y$ predicted by the model, we adopt the \emph{negative test log-likelihood} \cite{murphy2012machine}
\begin{equation}
    \text{NLL}(\mathcal{D}_{te},p)=-\frac{1}{|\mathcal{D}_{te}|}\sum_{(x,y)\in\mathcal{D}_{te}}\log(p(y|x)).
    \label{NNL_metric}
\end{equation} Note that the negative log-likelihood is large if the model assigns a small probability density $p(y|x)$ to the correct output $y$.



\subsection{Data Sets}
We experiment on different publicly available RSSI fingerprint data sets, encompassing both outdoor and indoor conditions:
\begin{itemize}
    \item The \textit{SigfoxRural} data set \cite{aernouts2018sigfox}  comprises $25,638$ Sigfox messages measured at 137 base stations and emitted from vehicles roaming around a large rural area ($1068\ \text{km}^2$) between Antwerp and Gent.
    \item The  \textit{UTSIndoorLoc} data set \cite{song2019novel} contains 9494 WiFi fingerprints sampled from 589 access points inside the FEIT Building at the University of Technology of Sydney, covering an area of $44,000\ \text{m}^2$.
    \item The  \textit{UJIIndoorLoc} data set \cite{torres2014ujiindoorloc} contains $21,049$ WiFi fingerprints measured at 520 access points and collected from 3 building of the Jaume I University, spanning a total area of $108,703\ \text{m}^2$.
  
\end{itemize}
To model the presence of \emph{outliers}, we modify the training data sets described above, producing $\epsilon$-contaminated data sets $\mathcal{D}$ as per Definition 2. This is done by replacing the target variable $y$ for a  fraction $\epsilon$ of the data points $(x,y)\in\mathcal{D}$ with a uniformly random location $y$ within the deployment area.


\subsection{Implementation}
We consider a model class specified by a Gaussian likelihood $p(y|x,\theta)=\mathcal{N}(y|f_\theta(x),0.01)$, where the mean $f_\theta(x)$ is the output of a neural network with two hidden layers, each  with 50 neurons with ELU activations. Despite the expressive power of the neural network model, each model $p(y|x,\theta)$ in this class can only account for unimodal, Gaussian distributed, residual uncertainties around the estimated position $f_\theta(x)$. Therefore, whenever the residual uncertainty about the receiver location is multimodal, the model class is \emph{misspecified} by Definition 1. As we will see, given the complex relation between RSSI vector and location, particularly when the number of RSSI measurements is sufficiently small, residual uncertainty tends to be multimodal, making this an important problem. 
Training for frequentist and Bayesian learning is carried out as described in the previous section, and ensembling uses $m=50$ samples during testing time.


\subsection{Results}

We start by considering the case in which there are no outliers, i.e., $\epsilon=0$, thus focusing solely on the problem of misspecification. In Figure \ref{fig:RSSI}, we plot the predictive distribution obtained via Bayesian learning ($m=1$, left panel) and robust Bayesian learning with $m=10$ and $t=1$ (right panel) for a testing sample $x$ corresponding to the position shown as a green cross. The black dots correspond to the positions covered by the training set in the  \textit{SigfoxRural} data set. The resulting predictive distribution for conventional Bayesian learning provides a poor estimation of the true device position, and is unable to properly quantify uncertainty. In contrast, robust Bayesian learning is able to counteract model misspecification, producing a more informative predictive distribution. The distribution correctly suggests that the receiver can be in two possible areas, one of which indeed containing the true node location. 

To further elaborate on the capacity of robust Bayesian learning for uncertainty quantification, in Table \ref{tab:KL} we report the negative log-likelihood (\ref{NNL_metric}) attained by Bayesian learning ($m=1$), as well as by robust Bayesian learning with $t=1$ and $m=2$ or $m=10$  on the three data sets. Increasing the value of $m$ is seen to yield lower negative log-likelihood scores, confirming that robust Bayesian learning provides a more precise quantification of uncertainty.


\begin{figure*}
\begin{subfigure}{0.33\textwidth}
   \includegraphics[width=1\linewidth]{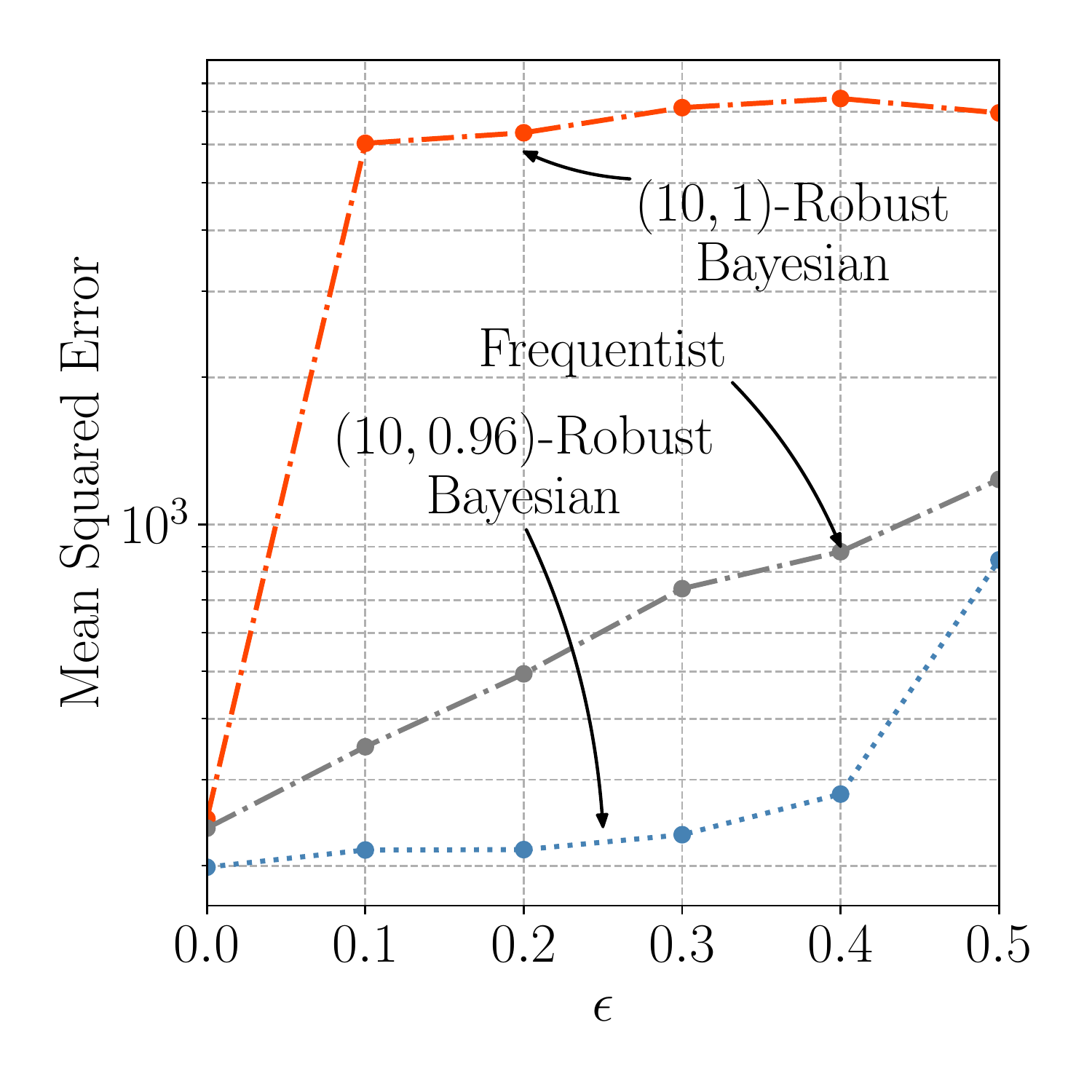}
   \caption{ \textit{SigfoxRural}}
   \label{fig:MSE1} 
\end{subfigure}
\hspace{-1.2em}
\begin{subfigure}{0.33\textwidth}
   \includegraphics[width=1\linewidth]{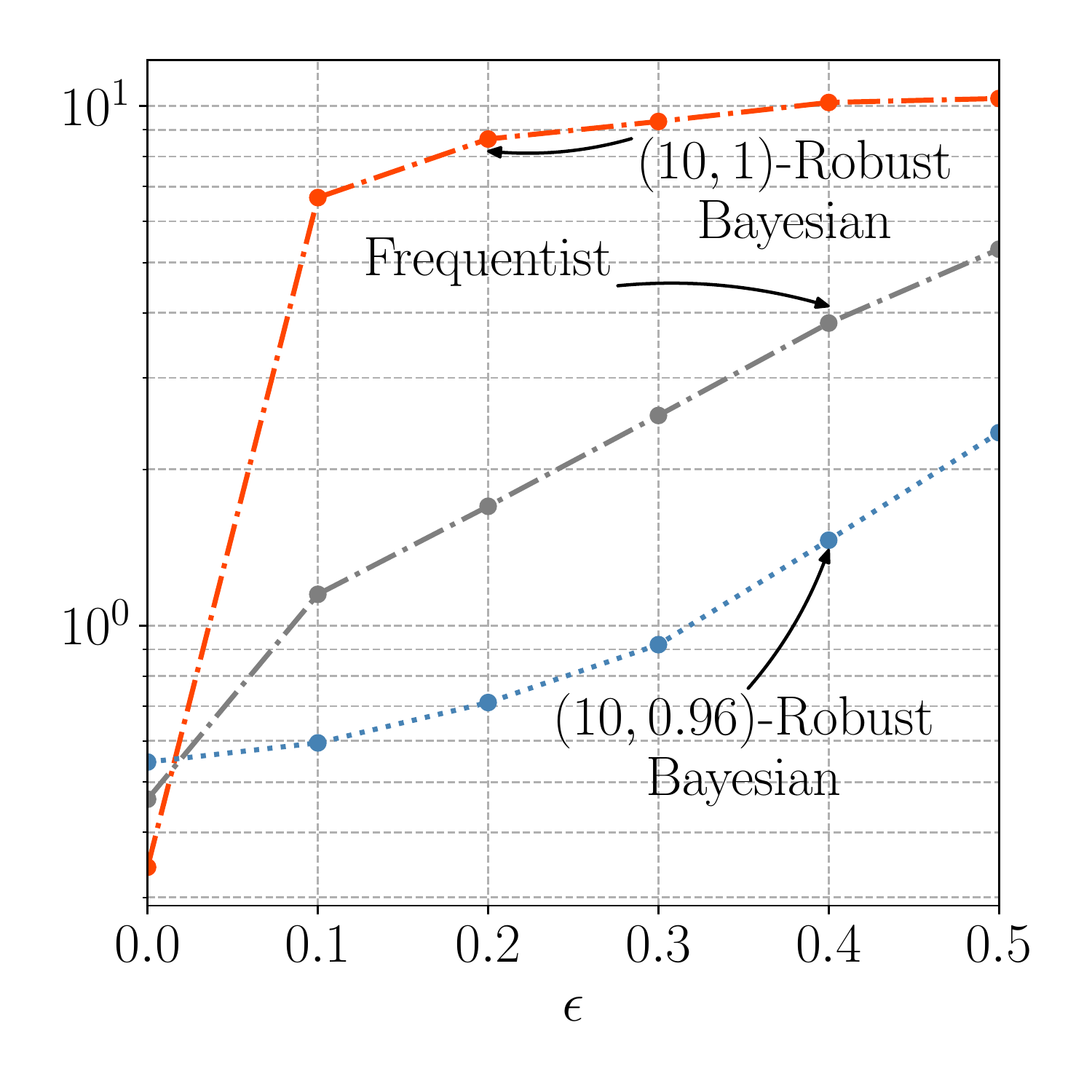}
    \caption{ \textit{UTSIndoor}}
   \label{fig:MSE2}
\end{subfigure}
\hspace{-1.2em}
\begin{subfigure}{0.33\textwidth}
   \includegraphics[width=1\linewidth]{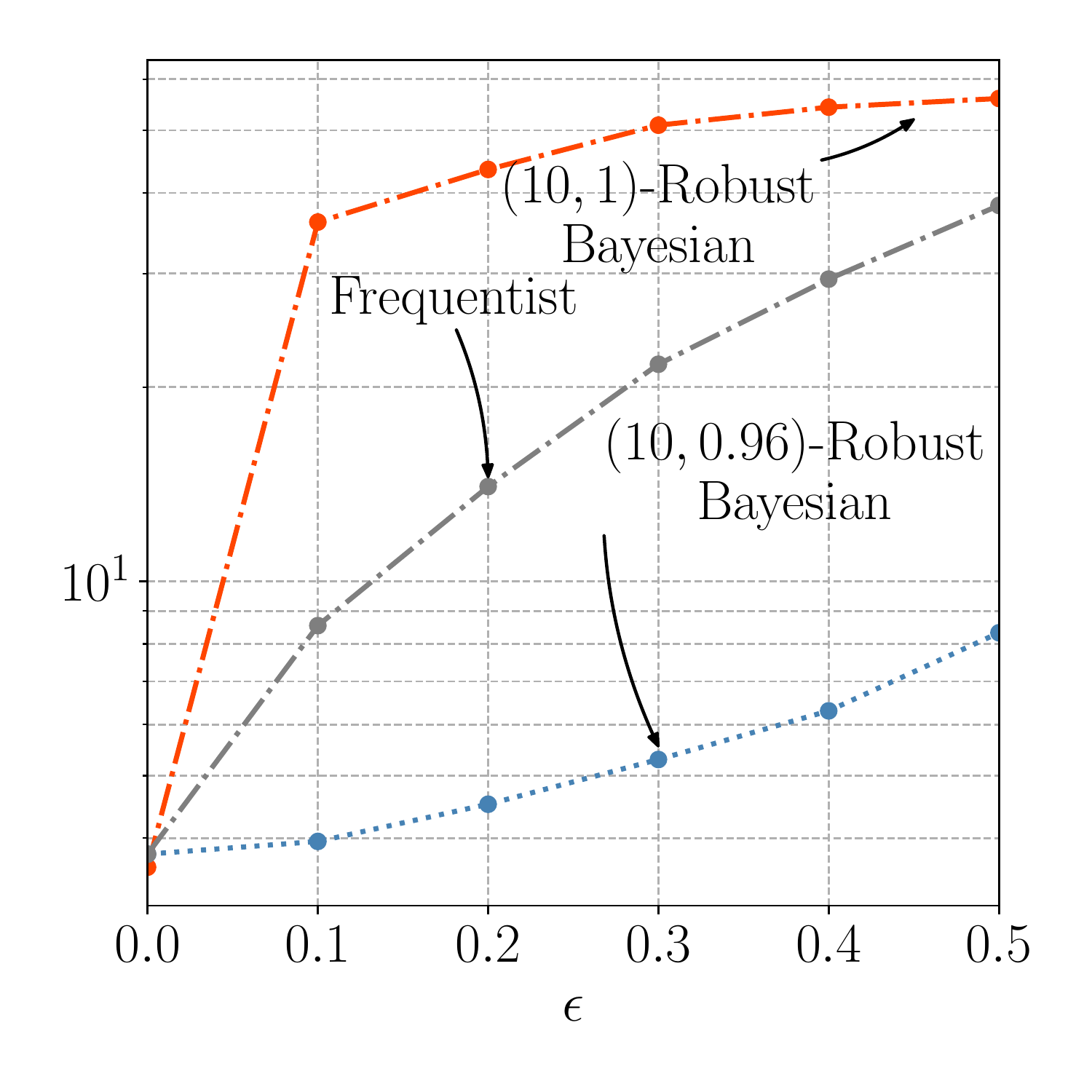}
    \caption{ \textit{UJIIndoor}}
   \label{fig:MSE2}
\end{subfigure}
\caption{Test mean squared error (\ref{MSE_metric}) for frequentist and the $(m,t)$-robust Bayesian learning with $m=10$ and $t=\{1,0.96\}$ as a  function of the corruption level $\epsilon$ for RSSI-based localization. As $\epsilon$ increases, the training data sets are increasingly affected by outliers.}
\label{fig:MSE}
\end{figure*}


We now introduce outliers by carrying out training on contaminated data sets with different levels of contamination $\epsilon$. Recall that the trained models are tested on a clean ($\epsilon=0$) test data set $\mathcal{D}_{te}$. In Figure \ref{fig:MSE}, we plot the test MSE (\ref{MSE_metric}) for frequentist and the $(m,t)$-robust Bayesian learning with $m=10$ and $t\in\{1,0.96\}$ as a function of $\epsilon$. The MSE of frequentist learning and $(10,1)$-robust Bayesian learning are seen to degrade significantly for increasing values of $\epsilon$. The performance loss is particularly severe for $(m,1)$-robust Bayesian learning. This is due to the mass-covering behavior entailed by the use of $m$-sample training loss, which in this case becomes detrimental due to the presence of outliers. In contrast, robust Bayesian learning with $t=0.96$ is able to counteract the effect of outliers, retaining good predictive performance even in case of largely corrupted data sets.

\section{Robust and Calibrated Channel Simulation}

\label{exp:ch_sim}
The design of communication systems has traditionally relied on analytical channel models obtained via measurements campaigns. Due to the complexity of multipath propagation scenarios, in recent years generative machine learning models have introduced as an alternative to analytical models. Generative models can be trained to produce samples that mimic hard-to-model channel conditions. Applications of deep generative models in the form of variational autoencoders (VAEs) \cite{kingma2013auto} and generative adversarial networks (GANs) \cite{goodfellow2014generative} were specifically reported in the context of end-to-end simulation of wireless systems in  \cite{aoudia2018end,ye2020deep} and for channel modeling in  \cite{o2019approximating,orekondy2022mimo,yang2019generative,ibnkahla2000applications} for earlier  applications to satellite communications.

The outlined prior work has focused on frequentist methods and has assumed the availability of clean data sets that are free from outliers. In this section, we explore the use of robust Bayesian learning to account for both outliers and model misspecification.


\subsection{Problem Definition and Performance Metrics} 

Generative models are trained in an unsupervised manner by assuming the availability of a training set $\mathcal{D}$ of examples $x$ corresponding to channel impulse responses. We focus on VAEs, i.e., on generative models with latent variables. VAEs  comprise a parameterized \emph{encoder} $q_{}(h|x,\theta_e)$, mapping an input $x\in\mathcal{X}$ into a lower-dimensional latent vector $h\in\mathcal{H}$; as well as a parameterized \emph{decoder} $p_{}(x|h,\theta_d)$ that reconstructs the input sample $x\in\mathcal{X}$ from the latent representation $h\in\mathcal{H}$. Note that the vector of model parameters encompasses both encoding and decoding parameters as  $\theta=(\theta_e,\theta_d)$. 


Let us define as $p(h)$ a fixed \emph{prior} distribution on the latent variables $h$. Once training is complete, samples $x$ of channel responses can be generated from the model as follows. For frequentist learning, given the trained model $\theta^{\textrm{freq}}$, one generates a sample $h \sim p(h)$ for the latent vector, and then produces a channel sample $x \sim p(x|h,\theta^{\textrm{freq}})$. For Bayesian learning, given the optimized distribution $q(\theta)$, we produce a random sample $\theta \sim q(\theta)$ and then generate channel sample $x \sim p(x|h,\theta_d)$. The role of the encoder $q_{}(h|x,\theta_e)$ will be made clear  in Section \ref{sec_impl_VAE} when discussing the training method.

According to the discussion in the previous paragraph, the channel distribution implemented by the model is given by 
\begin{align}
    p(x)=\mathbb{E}_{p(h)}[p(x|h,\theta^{\textrm{freq}}_d)]
\end{align}
for frequentist learning; and by 
\begin{align}
    p(x)=\mathbb{E}_{p(h)q(\theta_d)}[p(x|h,\theta_d)]
\end{align}
for Bayesian learning. Note that the average is taken only over the latent vector $h\sim p(h)$ for frequentist learning; while in Bayesian learning the expectation is also taken over the optimized distribution $q(\theta_d)$ for the decoder's parameters $\theta_d$.

To evaluate the performance of the generative model, we consider two different metrics accounting for accuracy and uncertainty quantification. Accuracy is measured by the ``distance'' between the target distribution $\nu(x)$ and the distribution $p_{}(x)$ produced by the model. 
We measure the ``distance" between $\nu(x)$ and $p(x)$ via the \emph{maximum-mean discrepancy} (MMD) \cite{gretton2006kernel}, which is defined as 
\begin{align}
    \mathrm{MMD}(p,\nu)=&\mathbb{E}_{x,x'\sim p(x)}[k(x,x')]+\mathbb{E}_{x,x'\sim \nu(x)}[k(x,x')]\nonumber\\&
    -2\mathbb{E}_{x\sim \nu(x),x'\sim p(x)}[k(x,x')]
\end{align}
where $k(x,x')$ is a positive definite kernel function. In the experiments reported below, we have approximated the MMD based on empirical averages. These are evaluated using samples from distribution $p(x)$, which are generated as explained above, as well as samples from the sampling distribution $\nu(x)$, i.e., examples from the training set $\mathcal{D}$. Moreover, we use the Gaussian kernel $k(x,x')=\mathcal{N}(\norm{x-x'}|0,1)$.


To evaluate the performance in terms of uncertainty quantification, we focus on the problem of \emph{out-of-distribution (OOD) detection} (see, e.g., \cite{daxberger2019bayesian}).
A well-calibrated model $p(x)$, when fed with an input $x$, should return a small value if $x$ is an OOD sample, that is, if it has a low target distribution $\nu(x)$.
To obtain a quantitative measure, we consider the task of distinguishing between samples drawn from the target distribution $\nu(x)$ and from the OOD distribution $\xi(x)$. Specifically, we adopt the  model probability distribution $p(x)$ as the test statistic, classifying $x$ as in-distribution (ID) if $p(x)$ is larger than some threshold $\gamma$ and as OOD otherwise. As in (\cite{fawcett2006introduction}), we take the area under the receiver operating characteristic curve (AUROC) score for this test as a measure of how distinguishable the two samples are. The AUROC metric is obtained by integrating the ROC traced by probability of detection versus probability of false alarm as the threshold $\gamma$ is varied. A larger AUROC indicates that the model provides a better quantification of uncertainty, as reflected in its capacity  to detect OOD samples against ID samples.

\begin{figure*}
\centering
\begin{subfigure}{0.45\textwidth}
   \includegraphics[width=1\linewidth]{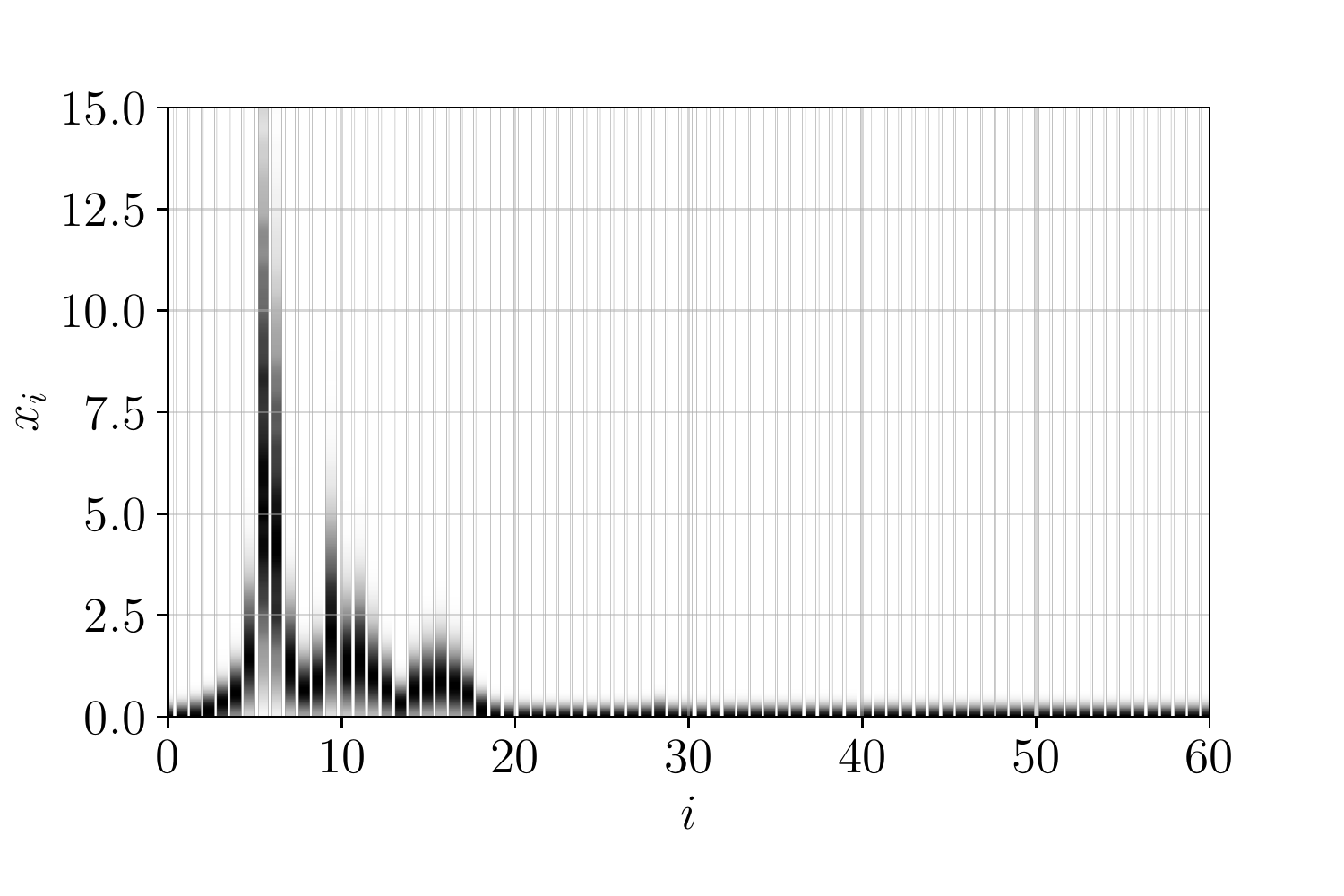}
   \caption{TDL-A $\tau=100$ns}
   \label{fig:TDLGT} 
\end{subfigure}
\begin{subfigure}{0.45\textwidth}
   \includegraphics[width=1\linewidth]{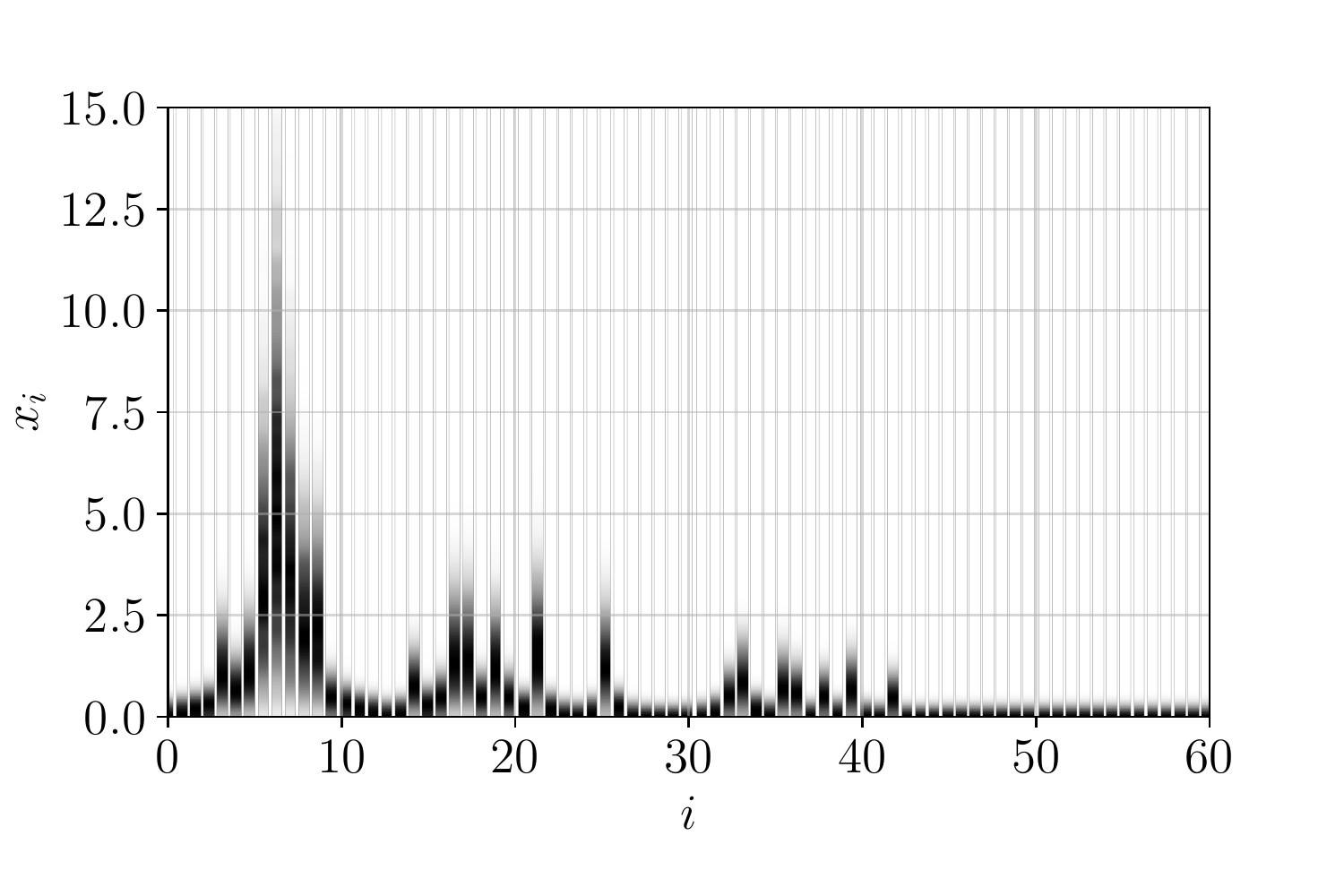}
   \caption{ TDL-A $\tau=300$ns}
   \label{fig:TDLOD} 
\end{subfigure}
\begin{subfigure}{0.45\textwidth}
   \includegraphics[width=1\linewidth]{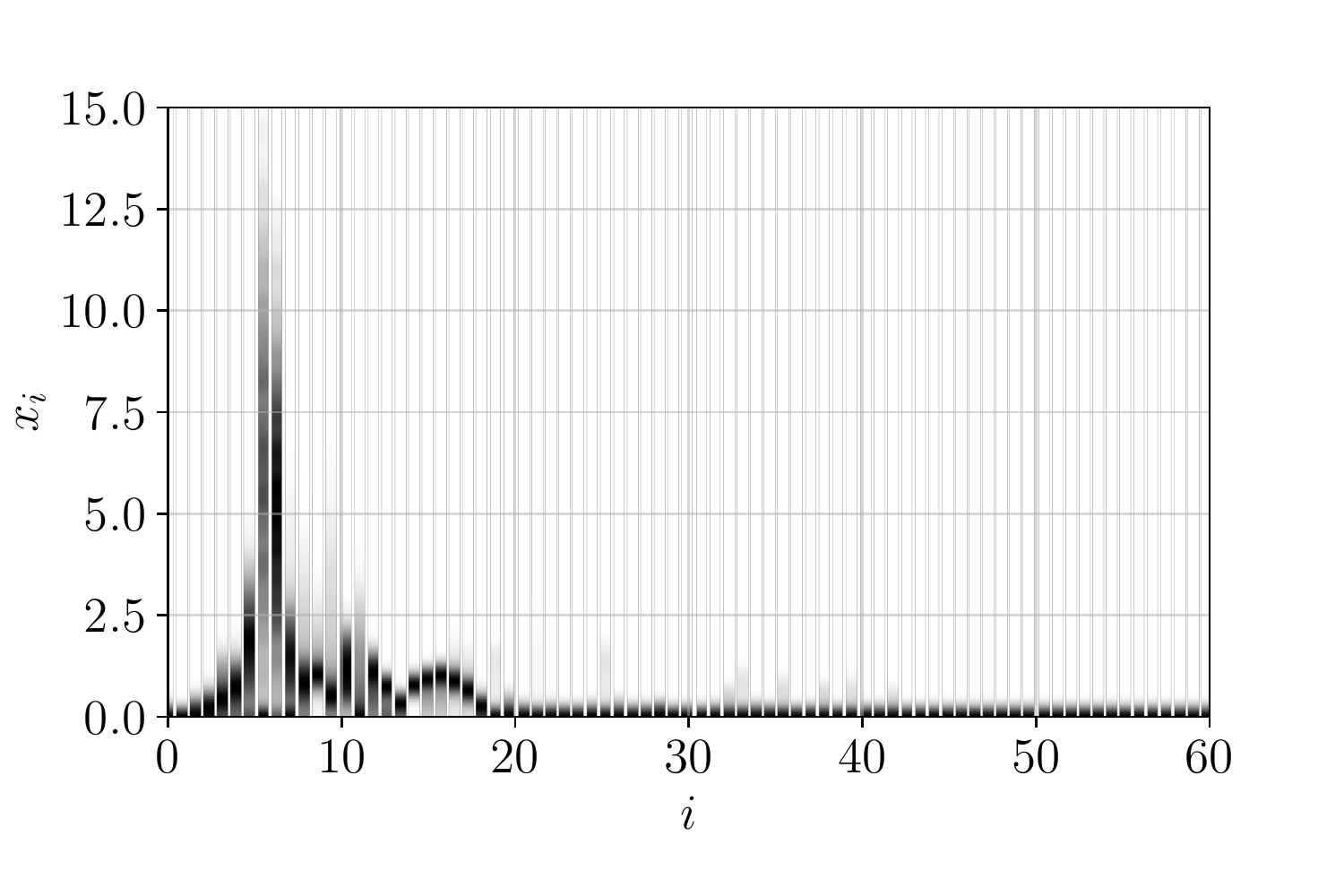}
    \caption{ Frequentist Learning}
   \label{fig:FreqVAE}
\end{subfigure}
\begin{subfigure}{0.45\textwidth}
   \includegraphics[width=1\linewidth]{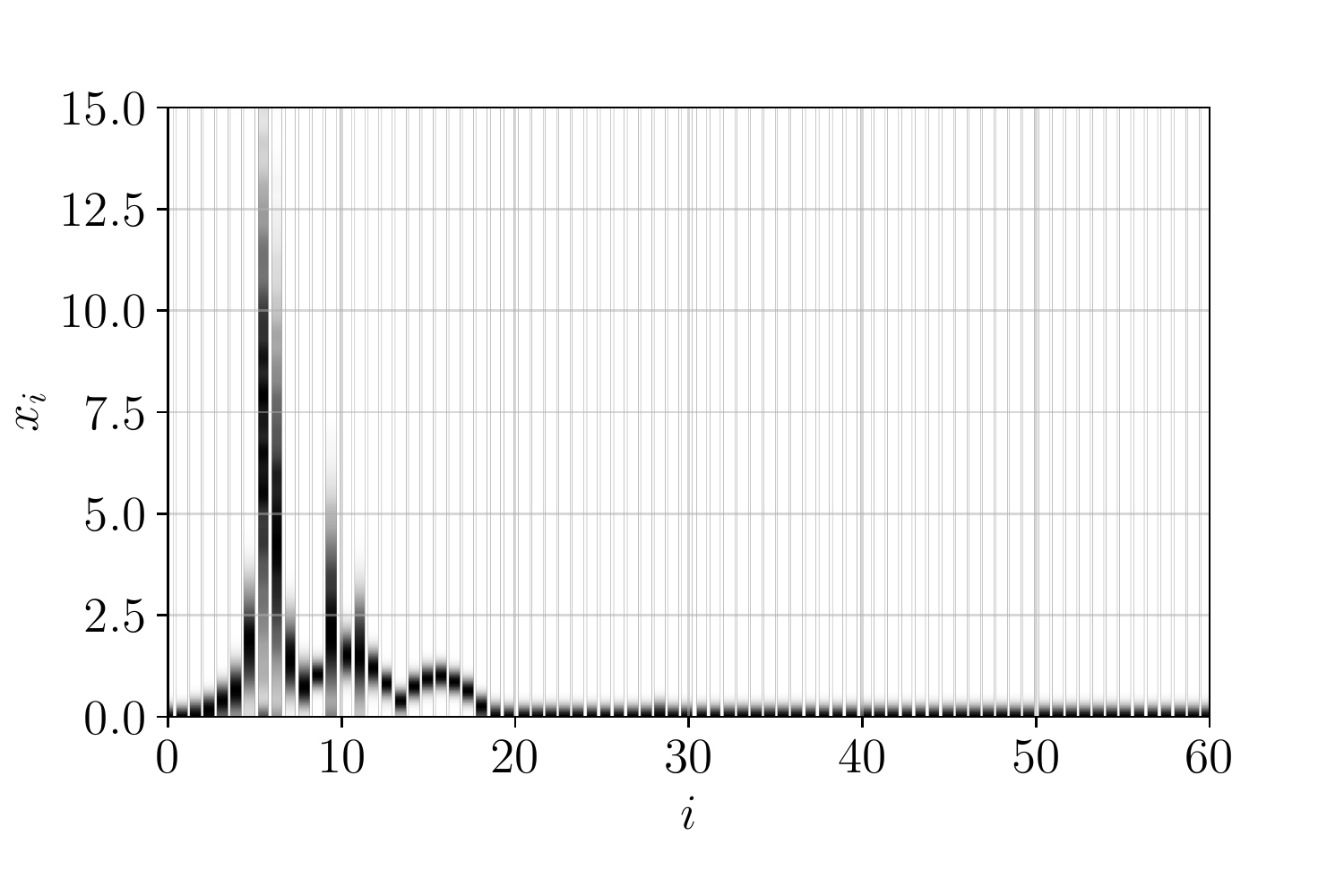}
    \caption{ $(4,0.7)$-Robust Bayesian Learning}
       \label{fig:BVAET0.7}
\end{subfigure}
\caption{\blue{The top row shows a sample of the magnitude for the TDL-A channel response given a delay spread $\tau=100$ns in panel (a), while an outlier sample corresponding to the larger delay spread $\tau=300$ ns is depicted in panel (b). The bottom row reports a sample from the trained model for frequentist learning in panel (c) and for $(4,0.7)$-robust Bayesian learning in panel (d).}}
\label{fig:CHDens}
\end{figure*}

\subsection{Data Set}
We consider the simulation of the magnitudes of a frequency-selective channel response $x\in\mathbb{R}^{128}$ that mimics the target distribution $\nu(x)$ defined by the 3GPP TDL-A channel model distribution \cite{etsi} with a delay spread of  $\tau=100$ ns. \emph{Outliers} are accounted for by constructing an $\epsilon$-contaminated training set $\mathcal{D}$ that contains a fraction $\epsilon=0.2$ of samples distributed according to the same  channel model but with a larger delay spread $\tau=300$ ns (see the top row in Fig. \ref{fig:CHDens}). 


\subsection{Implementation}
\label{sec_impl_VAE}
\blue{
For models with latent variables, the direct adoption of the log-loss generally yields intractable optimization problems (see, e.g., \cite{osvaldo2022ML4ENG}). To address this problem, training of VAEs replaces the training loss (\ref{emp_risk}) with the \emph{variational lower bound}
\begin{align}
   \mathcal{\hat{L}}^{VAE}(\theta,\mathcal{D})=&\sum_{x\in\mathcal{D}}\mathbb{E}_{p(h|x,\theta_e)}[\log(p(x|h,\theta_d)]\nonumber\\
   &-\sum_{x\in\mathcal{D}}\text{KL}(p(h|x,\theta_d)||p(h))\label{bound_likelihood},
\end{align}
which involves the use of the encoder model $p(h|x,\theta_e)$. 
Accordingly, the frequentist training objective is modified as 
\begin{align}
    \minimize_{\theta}\mathcal{\hat{L}}_{VAE}(\theta,\mathcal{D}),\label{vae_freq_obj}
\end{align}
while Bayesian learning addresses the problem
\begin{align}
    \minimize_{q(\theta)}\mathbb{E}_{q(\theta)}\left[\mathcal{\hat{L}}_{VAE}(\theta_e,\theta_d,\mathcal{D})\right]+\frac{1}{\beta}\text{KL}(q(\theta)||p(\theta)).\label{vae_freq_obj}
\end{align}
The robust free energy metrics are obtained in a similar manner, yielding the following formulation for  $(m,t)$-robust Bayesian learning
\begin{align}
   \hat{\mathcal{L}}^{VAE}_t\hspace{-0.1em}(\theta_1,\mydots,\theta_m,\mathcal{D})\hspace{-0.2em}=&\sum_{x\in\mathcal{D}}\mathbb{E}_{p(h|x,\theta_e)}\log_t\hspace{-0.2em}\left( \sum^m_{i=1} \frac{p(x|h,\theta_{d,i})}{m}\right)\nonumber\\
   &-\sum_{x\in\mathcal{D}}\text{KL}(p(h|x,\theta_d)||p(h))\label{bound_likelihood}.
\end{align}}

The prior latent variable distribution is $p(h)=\mathcal{N}(h|0,\mathbb{I}_5)$. We implement both the encoder and the decoder by using fully connected neural networks with a single hidden layer with 10 units. Specifically, the encoder distribution $p(h|x,\theta_e)=\mathcal{N}(h|\mu_{\theta_e}(x),\Sigma_{\theta_e}(x))$ has mean vector $\mu_{\theta_e}(x)\in\mathbb{R}^5$ and diagonal covariance matrix $\Sigma_{\theta_e}(x)\in\mathbb{R}^{5\times 5}$ obtained from the output of the neural network. The decoder  $p(x|h,\theta_d)=\mathcal{N}(\hat{x}|\mu_{\theta_d}(h),\sigma \mathbb{I}_{128})$ has mean vector $\mu_{\theta_d}(h)$ obtained as the output of the neural network with a fixed variance value $\sigma=0.1$. For Bayesian learning, we optimize distribution $q(\theta_d)$ as in the previous sections, while we consider a distribution $q(\theta_e)$ concentrated at a single vector $\theta_e$. Ensembling during testing time is carried out with $m=50$ samples.

\subsection{Results}

To start, in Figure \ref{fig:CHDens} we illustrate a sample of the magnitude for the TDL-A channel response given a delay spread $\tau=100$ ns in panel (a), while an outlier sample corresponding to the larger delay spread $\tau=300$ ns is depicted in panel (b). The bottom row of Figure \ref{fig:CHDens} reports a sample from the trained model for frequentist learning in panel (c) and for $(4,0.7)$-robust Bayesian learning in panel (d). Visual inspection of the last two panels confirms that $(m,t)$-robust Bayesian learning can mitigate the effect of outliers as it reduces the spurious multipath components associated with larger delays.

\begin{figure}
   \includegraphics[width=1\linewidth]{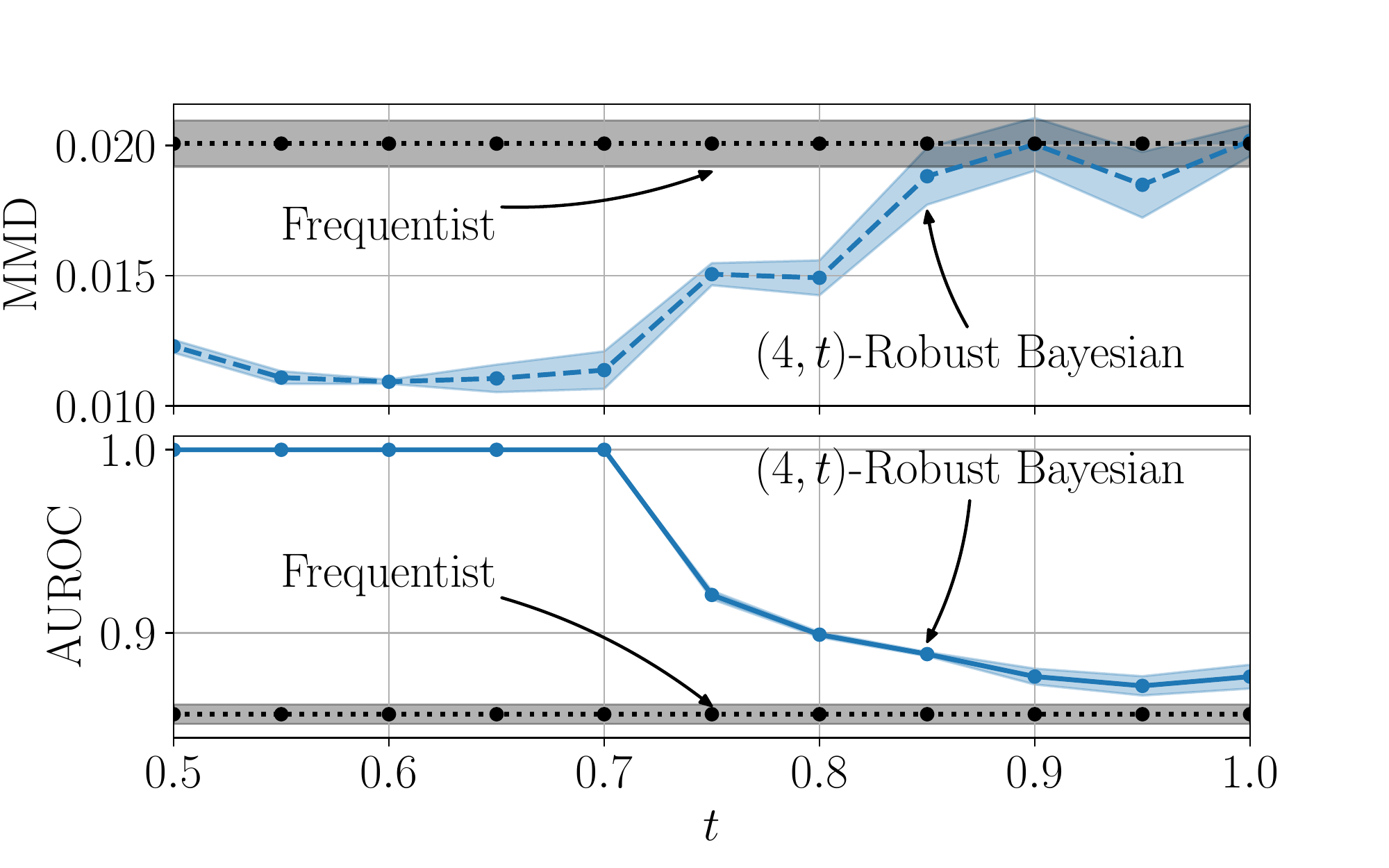}
   \caption{Maximum mean discrepancy (MMD) and area under receiving operating curve (AUROC) for frequentist learning and $(4,t)$-robust Bayesian learning. Both models are trained on a corrupted data set with $(\epsilon=0.2)$.  }
     \label{fig:MMD_AUROC_plot} 
\end{figure}

For a numerical comparison, Figure \ref{fig:MMD_AUROC_plot} compares frequentist and $(4,t)$-robust Bayesian learning in terms of both accuracy -- as measured by the MMD -- and uncertainty quantification -- as evaluated via the AUROC.  For  $t<0.85$ robust Bayesian learning is confirmed to have the capacity to mitigate the effect of the outlying component, almost halving the MMD obtained by frequentist learning. Furthermore, robust Bayesian learning has  a superior  uncertainty quantification performance, with gain increasing for decreasing values of $t$.

\section{Conclusion}
This work has focused on the problem of ensuring that AI models trained for wireless communications satisfy reliability and robustness requirements. We have specifically addressed two important problems: model misspecification, arising from limitations on the available knowledge about the problem and on the complexity of the AI models that can be implemented on network devices; and outliers, which cause a mismatch between training and testing conditions. We have argued that standard frequentist learning, as well as Bayesian learning, are not designed to address these requirements, and we have explored the application of \emph{robust Bayesian learning} to achieve robustness to model misspecification and to the presence of outliers in the training data set. Robust Bayesian learning has been shown to consistently provide better accuracy and uncertainty estimation capabilities in a range of important wireless communication problems.
These results motivate a range of extension of robust Bayesian learning and applications. For instance, the integration of robust Bayesian learning to the meta-learning framework, in order to enable robust and sample effective learning, or the application of robust Bayesian learning to higher layers of the protocol stack as a tool to empower semantic communication. 
\label{sec:concs}

\bibliographystyle{ieeetr}  
\bibliography{ref}

\begin{thebibliography}{10}

\bibitem{simeone2018very}
O.~Simeone, ``A very brief introduction to machine learning with applications
  to communication systems,'' {\em IEEE Transactions on Cognitive
  Communications and Networking}, vol.~4, no.~4, pp.~648--664, 2018.

\bibitem{sun2019application}
Y.~Sun, M.~Peng, Y.~Zhou, Y.~Huang, and S.~Mao, ``Application of machine
  learning in wireless networks: Key techniques and open issues,'' {\em IEEE
  Communications Surveys \& Tutorials}, vol.~21, no.~4, pp.~3072--3108, 2019.

\bibitem{gunduz2019machine}
D.~G{\"u}nd{\"u}z, P.~de~Kerret, N.~D. Sidiropoulos, D.~Gesbert, C.~R. Murthy,
  and M.~van~der Schaar, ``Machine learning in the air,'' {\em IEEE Journal on
  Selected Areas in Communications}, vol.~37, no.~10, pp.~2184--2199, 2019.

\bibitem{jiang2020learn}
Y.~Jiang, H.~Kim, H.~Asnani, S.~Kannan, S.~Oh, and P.~Viswanath, ``Learn codes:
  Inventing low-latency codes via recurrent neural networks,'' {\em IEEE
  Journal on Selected Areas in Information Theory}, vol.~1, no.~1,
  pp.~207--216, 2020.

\bibitem{kim2018deepcode}
H.~Kim, Y.~Jiang, S.~Kannan, S.~Oh, and P.~Viswanath, ``{Deepcode: Feedback
  codes via deep learning},'' {\em Advances in neural information processing
  systems}, vol.~31, 2018.

\bibitem{park2020meta}
S.~Park, O.~Simeone, and J.~Kang, ``{Meta-learning to communicate: Fast
  end-to-end training for fading channels},'' in {\em ICASSP 2020-2020 IEEE
  International Conference on Acoustics, Speech and Signal Processing
  (ICASSP)}, pp.~5075--5079, IEEE, 2020.

\bibitem{park2020learning}
S.~Park, H.~Jang, O.~Simeone, and J.~Kang, ``Learning to demodulate from few
  pilots via offline and online meta-learning,'' {\em IEEE Transactions on
  Signal Processing}, vol.~69, pp.~226--239, 2020.

\bibitem{simeone2020learning}
O.~Simeone, S.~Park, and J.~Kang, ``From learning to meta-learning: Reduced
  training overhead and complexity for communication systems,'' in {\em 2020
  2nd 6G Wireless Summit (6G SUMMIT)}, pp.~1--5, IEEE, 2020.

\bibitem{yuan2020transfer}
Y.~Yuan, G.~Zheng, K.-K. Wong, B.~Ottersten, and Z.-Q. Luo, ``{Transfer
  learning and meta learning-based fast downlink beamforming adaptation},''
  {\em IEEE Transactions on Wireless Communications}, vol.~20, no.~3,
  pp.~1742--1755, 2020.

\bibitem{guo2017calibration}
C.~Guo, G.~Pleiss, Y.~Sun, and K.~Q. Weinberger, ``On calibration of modern
  neural networks,'' in {\em International Conference on Machine Learning},
  pp.~1321--1330, PMLR, 2017.

\bibitem{cohen2021learning}
K.~M. Cohen, S.~Park, O.~Simeone, and S.~Shamai, ``Learning to learn to
  demodulate with uncertainty quantification via {B}ayesian meta-learning,''
  {\em arXiv preprint arXiv:2108.00785}, 2021.

\bibitem{masur2021artificial}
P.~H. Masur and J.~H. Reed, ``Artificial intelligence in {Open Radio Access
  Network},'' {\em arXiv preprint arXiv:2104.09445}, 2021.

\bibitem{mackay2003information}
D.~J.~C. MacKay, {\em Information theory, inference and learning algorithms}.
\newblock Cambridge University Press, 2003.

\bibitem{osawa2019practical}
K.~Osawa, S.~Swaroop, M.~E.~E. Khan, A.~Jain, R.~Eschenhagen, R.~E. Turner, and
  R.~Yokota, ``{Practical deep learning with Bayesian principles},'' {\em
  Advances in neural information processing systems}, vol.~32, 2019.

\bibitem{nikoloska2021bamld}
I.~Nikoloska and O.~Simeone, ``{BAMLD: Bayesian Active Meta-Learning by
  Disagreement},'' {\em arXiv preprint arXiv:2110.09943}, 2021.

\bibitem{madigan1996bayesian}
D.~Madigan, A.~E. Raftery, C.~Volinsky, and J.~Hoeting, ``{Bayesian model
  averaging},'' in {\em Proceedings of the AAAI Workshop on Integrating
  Multiple Learned Models, Portland, OR}, pp.~77--83, 1996.

\bibitem{osvaldo2022ML4ENG}
O.~Simeone, {\em {Machine Learning for Engineers}}.
\newblock Cambridge university press, 2022.

\bibitem{zilberstein2022annealed}
N.~Zilberstein, C.~Dick, R.~Doost-Mohammady, A.~Sabharwal, and S.~Segarra,
  ``Annealed {Langevin} dynamics for massive mimo detection,'' {\em arXiv
  preprint arXiv:2205.05776}, 2022.

\bibitem{masegosa2020learning}
A.~Masegosa, ``Learning under model misspecification: Applications to
  variational and ensemble methods,'' {\em Advances in Neural Information
  Processing Systems}, vol.~33, pp.~5479--5491, 2020.

\bibitem{morningstar2020pac}
W.~R. Morningstar, A.~A. Alemi, and J.~V. Dillon, ``{{PAC$^m$-B}ayes: narrowing
  the empirical risk gap in the misspecified Bayesian regime},'' {\em arXiv
  preprint arXiv:2010.09629}, 2020.

\bibitem{jose2021free}
S.~T. Jose and O.~Simeone, ``{Free energy minimization: A unified framework for
  modeling, inference, learning, and optimization [lecture notes]},'' {\em IEEE
  Signal Processing Magazine}, vol.~38, no.~2, pp.~120--125, 2021.

\bibitem{catoni2003pac}
O.~Catoni, ``{A {PAC-Bayesian} approach to adaptive classification},'' {\em
  preprint}, vol.~840, 2003.

\bibitem{alquier2021user}
P.~Alquier, ``{User-friendly introduction to PAC-Bayes bounds},'' {\em arXiv
  preprint arXiv:2110.11216}, 2021.

\bibitem{knoblauch2019generalized}
J.~Knoblauch, J.~Jewson, and T.~Damoulas, ``Generalized variational inference:
  Three arguments for deriving new posteriors,'' {\em arXiv preprint
  arXiv:1904.02063}, 2019.

\bibitem{basu1998robust}
A.~Basu, I.~R. Harris, N.~L. Hjort, and M.~Jones, ``{Robust and efficient
  estimation by minimising a density power divergence},'' {\em Biometrika},
  vol.~85, no.~3, pp.~549--559, 1998.

\bibitem{ghosh2016robust}
A.~Ghosh and A.~Basu, ``{Robust Bayes estimation using the density power
  divergence},'' {\em Annals of the Institute of Statistical Mathematics},
  vol.~68, no.~2, pp.~413--437, 2016.

\bibitem{fujisawa2008robust}
H.~Fujisawa and S.~Eguchi, ``{Robust parameter estimation with a small bias
  against heavy contamination},'' {\em Journal of Multivariate Analysis},
  vol.~99, no.~9, pp.~2053--2081, 2008.

\bibitem{nakagawa2020robust}
T.~Nakagawa and S.~Hashimoto, ``{Robust Bayesian inference via
  $\gamma$-divergence},'' {\em Communications in Statistics-Theory and
  Methods}, vol.~49, no.~2, pp.~343--360, 2020.

\bibitem{zecchin2022robust}
M.~Zecchin, S.~Park, O.~Simeone, M.~Kountouris, and D.~Gesbert, ``Robust
  {PAC$^m $}: Training ensemble models under model misspecification and
  outliers,'' {\em arXiv preprint arXiv:2203.01859}, 2022.

\bibitem{fawzy2013outliers}
A.~Fawzy, H.~M. Mokhtar, and O.~Hegazy, ``Outliers detection and classification
  in wireless sensor networks,'' {\em Egyptian Informatics Journal}, vol.~14,
  no.~2, pp.~157--164, 2013.

\bibitem{jin2015rssi}
R.~Jin, Z.~Che, H.~Xu, Z.~Wang, and L.~Wang, ``{An RSSI-based localization
  algorithm for outliers suppression in wireless sensor networks},'' {\em
  Wireless Networks}, vol.~21, no.~8, pp.~2561--2569, 2015.

\bibitem{kalyani2007ofdm}
S.~Kalyani and K.~Giridhar, ``{OFDM channel estimation in the presence of NBI
  and the effect of misspecified NBI model},'' in {\em 2007 IEEE 8th Workshop
  on Signal Processing Advances in Wireless Communications}, pp.~1--5, IEEE,
  2007.

\bibitem{liang2011cognitive}
Y.-C. Liang, K.-C. Chen, G.~Y. Li, and P.~Mahonen, ``Cognitive radio networking
  and communications: An overview,'' {\em IEEE transactions on vehicular
  technology}, vol.~60, no.~7, pp.~3386--3407, 2011.

\bibitem{lohan2017wi}
E.~S. Lohan, J.~Torres-Sospedra, H.~Lepp{\"a}koski, P.~Richter, Z.~Peng, and
  J.~Huerta, ``{Wi-Fi} crowdsourced fingerprinting dataset for indoor
  positioning,'' {\em Data}, vol.~2, no.~4, p.~32, 2017.

\bibitem{blei2017variational}
D.~M. Blei, A.~Kucukelbir, and J.~D. McAuliffe, ``Variational inference: A
  review for statisticians,'' {\em Journal of the American Statistical
  Association}, vol.~112, no.~518, pp.~859--877, 2017.

\bibitem{xiao2013identification}
Z.~Xiao, H.~Wen, A.~Markham, N.~Trigoni, P.~Blunsom, and J.~Frolik,
  ``Identification and mitigation of non-line-of-sight conditions using
  received signal strength,'' in {\em 2013 IEEE 9th International Conference on
  Wireless and Mobile Computing, Networking and Communications (WiMob)},
  pp.~667--674, IEEE, 2013.

\bibitem{catoni2007pac}
O.~Catoni, ``{PAC}-bayesian supervised classification: the thermodynamics of
  statistical learning,'' {\em arXiv preprint arXiv:0712.0248}, 2007.

\bibitem{Huber}
P.~J. Huber, ``{Robust estimation of a location parameter},'' {\em The Annals
  of Mathematical Statistics}, vol.~35, no.~1, pp.~73--101, 1964.

\bibitem{jewson2018principles}
J.~Jewson, J.~Q. Smith, and C.~Holmes, ``{Principles of Bayesian inference
  using general divergence criteria},'' {\em Entropy}, vol.~20, no.~6, p.~442,
  2018.

\bibitem{o2018over}
T.~J. O’Shea, T.~Roy, and T.~C. Clancy, ``Over-the-air deep learning based
  radio signal classification,'' {\em IEEE Journal of Selected Topics in Signal
  Processing}, vol.~12, no.~1, pp.~168--179, 2018.

\bibitem{o2016convolutional}
T.~J. O’Shea, J.~Corgan, and T.~C. Clancy, ``Convolutional radio modulation
  recognition networks,'' in {\em International conference on engineering
  applications of neural networks}, pp.~213--226, Springer, 2016.

\bibitem{zhou2020deep}
R.~Zhou, F.~Liu, and C.~W. Gravelle, ``Deep learning for modulation
  recognition: A survey with a demonstration,'' {\em IEEE Access}, vol.~8,
  pp.~67366--67376, 2020.

\bibitem{degroot1983comparison}
M.~H. DeGroot and S.~E. Fienberg, ``The comparison and evaluation of
  forecasters,'' {\em Journal of the Royal Statistical Society: Series D (The
  Statistician)}, vol.~32, no.~1-2, pp.~12--22, 1983.

\bibitem{kingma2013auto}
D.~P. Kingma and M.~Welling, ``Auto-encoding variational bayes,'' {\em arXiv
  preprint arXiv:1312.6114}, 2013.

\bibitem{mautz2009overview}
R.~Mautz, ``Overview of current indoor positioning systems,'' {\em Geodezija ir
  kartografija}, vol.~35, no.~1, pp.~18--22, 2009.

\bibitem{aernouts2018sigfox}
M.~Aernouts, R.~Berkvens, K.~Van~Vlaenderen, and M.~Weyn, ``Sigfox and
  {LoRaWAN} datasets for fingerprint localization in large urban and rural
  areas,'' {\em Data}, vol.~3, no.~2, p.~13, 2018.

\bibitem{pecoraro2018csi}
G.~Pecoraro, S.~Di~Domenico, E.~Cianca, and M.~De~Sanctis, ``{CSI-based
  fingerprinting for indoor localization using LTE signals},'' {\em EURASIP
  Journal on Advances in Signal Processing}, vol.~2018, no.~1, pp.~1--18, 2018.

\bibitem{hoang2019recurrent}
M.~T. Hoang, B.~Yuen, X.~Dong, T.~Lu, R.~Westendorp, and K.~Reddy, ``{Recurrent
  neural networks for accurate RSSI indoor localization},'' {\em IEEE Internet
  of Things Journal}, vol.~6, no.~6, pp.~10639--10651, 2019.

\bibitem{sinha2019comparison}
R.~S. Sinha and S.-H. Hwang, ``{Comparison of CNN applications for RSSI-based
  fingerprint indoor localization},'' {\em Electronics}, vol.~8, no.~9, p.~989,
  2019.

\bibitem{murphy2012machine}
K.~P. Murphy, {\em Machine learning: a probabilistic perspective}.
\newblock MIT press, 2012.

\bibitem{song2019novel}
X.~Song, X.~Fan, C.~Xiang, Q.~Ye, L.~Liu, Z.~Wang, X.~He, N.~Yang, and G.~Fang,
  ``A novel convolutional neural network based indoor localization framework
  with {WiFi} fingerprinting,'' {\em IEEE Access}, vol.~7, pp.~110698--110709,
  2019.

\bibitem{torres2014ujiindoorloc}
J.~Torres-Sospedra, R.~Montoliu, A.~Mart{\'\i}nez-Us{\'o}, J.~P. Avariento,
  T.~J. Arnau, M.~Benedito-Bordonau, and J.~Huerta, ``Ujiindoorloc: A new
  multi-building and multi-floor database for {WLAN} fingerprint-based indoor
  localization problems,'' in {\em 2014 international conference on indoor
  positioning and indoor navigation (IPIN)}, pp.~261--270, IEEE, 2014.

\bibitem{goodfellow2014generative}
I.~Goodfellow, J.~Pouget-Abadie, M.~Mirza, B.~Xu, D.~Warde-Farley, S.~Ozair,
  A.~Courville, and Y.~Bengio, ``Generative adversarial nets,'' {\em Advances
  in neural information processing systems}, vol.~27, 2014.

\bibitem{aoudia2018end}
F.~A. Aoudia and J.~Hoydis, ``End-to-end learning of communications systems
  without a channel model,'' in {\em 2018 52nd Asilomar Conference on Signals,
  Systems, and Computers}, pp.~298--303, IEEE, 2018.

\bibitem{ye2020deep}
H.~Ye, L.~Liang, G.~Y. Li, and B.-H. Juang, ``Deep learning-based end-to-end
  wireless communication systems with conditional {GANs} as unknown channels,''
  {\em IEEE Transactions on Wireless Communications}, vol.~19, no.~5,
  pp.~3133--3143, 2020.

\bibitem{o2019approximating}
T.~J. O’Shea, T.~Roy, and N.~West, ``Approximating the void: Learning
  stochastic channel models from observation with variational generative
  adversarial networks,'' in {\em 2019 International Conference on Computing,
  Networking and Communications (ICNC)}, pp.~681--686, IEEE, 2019.

\bibitem{orekondy2022mimo}
T.~Orekondy, A.~Behboodi, and J.~B. Soriaga, ``{MIMO-GAN}: Generative {MIMO}
  channel modeling,'' {\em arXiv preprint arXiv:2203.08588}, 2022.

\bibitem{yang2019generative}
Y.~Yang, Y.~Li, W.~Zhang, F.~Qin, P.~Zhu, and C.-X. Wang,
  ``Generative-adversarial-network-based wireless channel modeling: Challenges
  and opportunities,'' {\em IEEE Communications Magazine}, vol.~57, no.~3,
  pp.~22--27, 2019.

\bibitem{ibnkahla2000applications}
M.~Ibnkahla, ``Applications of neural networks to digital communications--a
  survey,'' {\em Signal processing}, vol.~80, no.~7, pp.~1185--1215, 2000.

\bibitem{gretton2006kernel}
A.~Gretton, K.~Borgwardt, M.~Rasch, B.~Sch{\"o}lkopf, and A.~Smola, ``A kernel
  method for the two-sample-problem,'' {\em Advances in neural information
  processing systems}, vol.~19, 2006.

\bibitem{daxberger2019bayesian}
E.~Daxberger and J.~M. Hern{\'a}ndez-Lobato, ``Bayesian variational
  autoencoders for unsupervised out-of-distribution detection,'' {\em arXiv
  preprint arXiv:1912.05651}, 2019.

\bibitem{fawcett2006introduction}
T.~Fawcett, ``An introduction to {ROC} analysis,'' {\em Pattern recognition
  letters}, vol.~27, no.~8, pp.~861--874, 2006.

\bibitem{etsi}
D.~3rd Generation Partnership Project~(3GPP), ``Study on channel model for
  frequencies from 0.5 to 100 {GHz},'' {\em 3GPP TR 38.901}, 2020.

\end{thebibliography}

\end{document}